\documentclass{article} 
\usepackage{iclr2026_conference,times}

\usepackage{amsmath,amsfonts,bm}

\def\eqref#1{equation~\ref{#1}}

\def\1{\bm{1}}

\DeclareMathAlphabet{\mathsfit}{\encodingdefault}{\sfdefault}{m}{sl}
\SetMathAlphabet{\mathsfit}{bold}{\encodingdefault}{\sfdefault}{bx}{n}

\usepackage{hyperref}
\usepackage{url}
\usepackage[table]{xcolor}
\usepackage{array}
\usepackage{microtype}
\usepackage{graphicx}
\usepackage{subcaption}
\usepackage{booktabs}
\usepackage{amsmath}
\usepackage{amssymb}
\usepackage{mathtools}
\usepackage{amsthm}
\usepackage{times}
\usepackage{latexsym}
\usepackage[T1]{fontenc}
\usepackage[utf8]{inputenc}
\usepackage{graphicx}
\usepackage{xspace}
\usepackage{tabularx}
\usepackage{wrapfig}
\usepackage{multirow}
\usepackage{graphicx}
\usepackage{pifont}
\usepackage{float}
\usepackage{enumitem}
\usepackage{times}
\usepackage{latexsym}
\usepackage{booktabs}
\usepackage{arydshln}
\usepackage{makecell}
\usepackage{comment}
\usepackage{cleveref}
\usepackage{placeins}
\usepackage{stfloats}
\usepackage[flushmargin]{footmisc}

\title{Preference-Aware Rubric Learning for Personalized Evaluation}

\author{
\textbf{Yilun Qiu\textsuperscript{1,2}} 
~~
\textbf{Xiaoyan Zhao\textsuperscript{1*}}
~~
\textbf{Yang Zhang\textsuperscript{1*}}
~~
\textbf{Yuxin Chen\textsuperscript{1}}
~~
\textbf{Cilin Yan\textsuperscript{2}}
\\\
\textbf{Jiayin Cai\textsuperscript{2}} 
~~
\textbf{Xiaolong Jiang\textsuperscript{2}}
~~
\textbf{Yao Hu\textsuperscript{2}}
~~
\textbf{Yoko Yamakata\textsuperscript{3}}
~~
\textbf{Tat-Seng Chua\textsuperscript{1}}
\\
\textsuperscript{1}National University of Singapore
~~
\textsuperscript{2}Xiaohongshu Inc.
~~
\textsuperscript{3}The University of Tokyo
\\
}

\newcommand{\eg}{\emph{e.g., }}

\newcolumntype{C}{>{\centering\arraybackslash}X}

\newcommand{\ours}{PARL\xspace}
\newcommand{\oursfull}{\textbf{P}reference-\textbf{A}ware \textbf{R}ubric \textbf{L}earning\xspace}

\iclrfinalcopy
\begin{document}

\maketitle

\begingroup
\renewcommand{\thefootnote}{\fnsymbol{footnote}}
\footnotetext[1]{Corresponding Authors}
\endgroup

\begin{abstract}
As Large Language Models (LLMs) evolve from general-purpose assistants to user-centric agents, personalization has become central to aligning model behavior with individual preferences, making the evaluation of personalized alignment a critical bottleneck. Existing evaluation methods—ranging from automatic metrics to LLM-as-a-judge approaches—fail to capture subjective, user-specific preferences embedded in long-term interaction histories. We identify three essential principles for reliable and effective personalized evaluation: \emph{Representativeness}, \emph{User-Consistency}, and \emph{Discriminativeness}.
To address these principles, we introduce \textit{Personalized Evaluation as Learning}, a paradigm that formulates personalized evaluation as a learning problem rather than a static judgment. Under this paradigm, we propose \textbf{\ours}~(\oursfull), a framework that learns to induce preference-aware evaluation rubrics directly from raw user histories and performs a self-validation mechanism to ensure consistency with the user’s preferences. \ours integrates rubric induction with a discriminative reinforcement learning objective that contrasts user-authored responses against competitive personalized model outputs, enabling the learned rubrics to capture precise, user-specific decision boundaries.
Experiments on real-world personalized text generation tasks show that \ours consistently induces high-fidelity rubrics that reliably identify user-aligned responses and generalize across users and tasks, while capturing stable stylistic preferences and fine-grained evaluative patterns.
To ensure reproducibility, our code is available at \url{https://github.com/SnowCharmQ/PARL}.
\end{abstract}

\section{Introduction}

Large Language Models (LLMs)~\citep{gpt4,deepseekr1,zhang2025collm} have significantly advanced general-purpose intelligence and are now widely applied to reasoning, content generation, and decision-making tasks. However, their prevailing \textit{``one-size-fits-all''} paradigm limits their ability to accommodate diverse and user-specific preferences. This limitation has motivated \textit{LLM Personalization}~\citep{personalizationsurvey1,personalizationsurvey2}, which aims to adapt model behavior to individual users across personalized tasks such as writing assistance and conversational interaction. Currently, LLM personalization has become an important and rapidly growing research direction, attracting increasing attention from both academia~\citep{nextquill,synthesizeme,flythinker} and industry~\citep{personamemv1,prefeval,vitabench2}.

Existing research on LLM personalization has primarily focused on aligning model behavior with individual user preferences through improvements to the \emph{generation process}, leading to a range of approaches such as user-specific context injection~\citep{dpl,lamp,hydra} and fine-tuning on historical user contexts~\citep{oppu,perfit,nextquill}.
In contrast, the \emph{evaluation} of LLM personalization remains largely underexplored, despite being a critical bottleneck for personalization method design and verification. This gap stems from a fundamental mismatch in evaluation assumptions: conventional evaluation methods rely on universal and objective ground truths, whereas personalized generation is inherently governed by subjective, user-specific criteria that are usually unobservable and only implicitly reflected in users’ historical behaviors.

Existing evaluation methods for LLM personalization can be broadly grouped into three categories: (1) \textit{Automatic metrics}, like ROUGE~\citep{rouge} and BERTScore~\citep{bertscore}. These metrics generally assume a single objective reference and prioritize surface-level lexical overlap to the reference. As a result, they are inherently inadequate for personalized evaluation, failing to capture subjective, user-specific preferences and alternative but equally valid personalized responses.
(2) \textit{Human evaluation}, which is often regarded as the gold standard in general scenarios. Yet, it suffers from an inferential gap in personalization settings: external annotators lack direct access to a user’s true preferences, making it difficult to faithfully assess alignment within the user. 
(3) \textit{LLM-as-a-judge} approaches~\citep{geval,gptscore}, which partially alleviate these issues via LLMs' semantic capabilities. However, they rely on static judges and handcrafted prompts, restricting adaptation to individual users and task-specific preferences, and they typically yield coarse, holistic scores without fine-grained grounding, limiting the interpretability and verifiability of their judgments.



Given the limitations of existing approaches, we propose a shift toward a \textit{Personalized Evaluation as Learning} paradigm for LLM personalization.
Rather than relying on static and generic evaluation rules, this paradigm aims to adaptively and systematically learn user-specific evaluation criteria aligned with individual preferences.
To realize this paradigm, we build upon the LLM-as-judge framework but depart from existing practices by instantiating it through \emph{rubric-based evaluation}~\citep{prometheus,llmrubric,pathak2025rubric}, which decomposes holistic judgments into explicit, multi-dimensional criteria, referred to as rubrics.
Such a formulation provides a natural foundation for learning-based personalized evaluation:
rubrics can be adaptively induced from individual user historical contexts and applied across diverse scenarios.
The core challenge, thereafter, is to learn how to drive rubrics that faithfully capture individual user-specific decision boundaries.


To answer this, we identify and formalize three essential principles that any rigorous and reliable rubric-based framework for personalized evaluation must satisfy:
(i) \textbf{\textit{Representativeness}}: Rubrics should faithfully distill the diverse and latent preferences embedded in a user’s historical contexts, capturing the key dimensions that define their evaluative standards.
(ii) \textbf{\textit{User-Consistency}}: Rubrics should reflect stable, user-specific preferences that persist across different historical contexts, rather than artifacts arising from isolated tasks or individual samples.
(iii) \textbf{\textit{Discriminativeness}}: Rubrics should enable fine-grained distinction between authentic user-authored responses and generic yet high-quality model outputs, isolating true preference awareness from universal quality signals.


Guided by the three principles, we propose \textbf{\ours} (\oursfull), a framework that learns to induce structured, multi-dimensional, preference-aware rubrics from raw user historical behaviors to support personalized LLM evaluation.
\ours distills semantically rich, user-specific preferences from users' historical contexts into explicit rubric dimensions, ensuring \emph{representativeness}. 
The induced rubrics are then validated across diverse historical contexts to enforce \emph{user-consistency}.
To further enhance \emph{discriminativeness}, \ours incorporates a reinforcement learning stage that contrasts user-authored ground-truth responses with competitive personalized model outputs, encouraging learned rubrics to capture precise, user-specific decision boundaries.


We conduct comprehensive experiments on real-world personalized text generation tasks to systematically assess the learned rubrics.
Empirical results demonstrate that \ours consistently induces rubrics that are representative, user-consistent, and discriminative. Further analysis reveals that the learned rubrics capture stable stylistic preferences and fine-grained criteria, reflecting distinct evaluative signatures.
Moreover, the trained rubric generator generalizes well across contexts, inducing rubrics that are semantically rich, diverse, and tailored to individual users.
Remarkably, these rubrics constitute a novel, reliable, and high-fidelity set of personalized evaluation metrics.

Our main contributions are summarized as follows:
\begin{itemize}[leftmargin=*, topsep=2pt, itemsep=0pt]
    \item We introduce \textit{Personalized Evaluation as Learning} paradigm, which formalizes personalized LLM evaluation as a learnable process grounded in three essential criteria: \textit{Representativeness}, \textit{User-Consistency}, and \textit{Discriminativeness}.
    \item We propose \textbf{\ours},
    a learning-based framework that induces preference-aware evaluation rubrics from user histories, optimized with a discriminative margin objective via RL.
    \item We conduct extensive experiments and analysis across real-world personalized generation tasks, demonstrating that \ours produces high-fidelity, reusable rubrics and offering new insights into the structure of user-specific evaluative preferences.
\end{itemize}

\section{Related Work}

In this section, we review related research across two dimensions: the development of LLM personalization and the evaluation of personalized text generation.

\subsection{LLM Personalization}

The widespread use of LLMs across diverse domains~\citep{llm2rec,selfverify,zhang2026reinforced,nextmem} has catalyzed a paradigm shift from \textit{``one-size-fits-all''} alignment toward personalized outputs tailored to individual preferences.
Current research in LLM personalization primarily falls into two tracks: personalized generation~\citep{personalizationsurvey3} and personalized preference alignment~\citep{personalizationsurvey6,zhang2024causality}.

Personalized generation has been studied across modalities such as text, image, and audio~\citep{learning,pigeon,vibemus}, with personalized text generation serving as a foundational direction~\citep{lamp,steerx,personalizationsurvey1}.
LaMP~\citep{lamp} and LongLaMP~\citep{longlamp} establish benchmarks for short- and long-form personalized generation, respectively.
Existing methods explore diverse strategies to leverage user historical context for personalized generation; for example, HYDRA~\citep{hydra} employs personalized reranking to prioritize the most relevant user histories during retrieval, while DEP~\citep{dep} distills inter-user differences into soft prompts for fine-grained guidance.

For personalized preference alignment, existing work primarily focuses on modeling personalized reward functions~\citep{psoups,pad,micro}. Representative approaches decompose rewards into shared and user-specific components~\citep{pal}, infer user contexts through latent variables~\citep{vpl}, or constrain personalized rewards within low-dimensional subspaces~\citep{lore,pref1}. P-GenRM~\citep{pgenrm} further adopts a multi-stage reinforcement learning framework to learn personalized reward models.
In contrast, our work shifts the focus from learning personalized reward functions to \emph{evaluating} personalized text generation. Rather than producing implicit scalar rewards, we aim to induce explicit, structured evaluation rubrics that serve as reliable benchmarks for personalized alignment. Our framework grounds rubric induction in user histories and enforces representativeness, user-consistency, and discriminativeness, providing principled evaluative guidance that is absent in existing personalized reward modeling approaches.

\subsection{Evaluation of Personalized Text Generation}

Existing evaluation methods for personalized text generation generally fall into three paradigms: automatic metrics, human evaluation, and LLM-based evaluation.
Automatic metrics, including $n$-gram overlap measures such as ROUGE, BLEU, and METEOR~\citep{rouge,bleu,meteor}, provide simple and efficient baselines but rely on surface-level lexical matching and assume a single objective reference. 
Even semantic-aware extensions such as BERTScore~\citep{bertscore} and BARTScore~\citep{bartscore}, which compare representations in high-dimensional embedding spaces, remain fundamentally limited for personalization, as they cannot capture subjective, user-specific preferences or multiple valid personalized responses.
Human evaluation is often regarded as the gold standard for qualitative assessment. 
However, in personalized settings, it suffers from an inherent inferential gap: external annotators lack direct access to users’ latent preferences, making it difficult to faithfully assess alignment with a target user.

LLM-based evaluation~\citep{chatbotarena,prometheus,llmasjudgesurvey,cce}, leverages the semantic and linguistic capabilities of large models to approximate human judgments. While effective for general-purpose evaluation, these approaches remain constrained by static pre-trained judges and handcrafted prompts, which are neither adaptive to individual users nor sensitive to task-specific preference variations. Moreover, they typically produce coarse, holistic scores without explicit, fine-grained grounding, limiting interpretability and verifiability.
Recent work has explored rubric-based evaluation, which guides assessment using explicit, multi-dimensional criteria~\citep{llmrubric,rubicon,openrubrics,rulers}. Several studies have further applied LLM-based judges or structured evaluation frameworks to personalized generation tasks~\citep{dpl,expert,restpg,lampqa,aupel,pref2}. However, existing approaches largely rely on static or heuristically induced rubrics, lack principled grounding in user histories, and provide limited analysis of rubric quality. These limitations motivate the need for evaluation frameworks that are explicitly grounded in user behavioral histories and capable of learning representative, consistent, and discriminative criteria for personalized text generation- this is the goal of our method.

\section{Preliminary}

In this section, we formally define the problem of personalized evaluation and introduce the rubric-based evaluation paradigm that serves as the foundation for our proposed framework.

\subsection{Problem Formulation}

Let $\mathcal{U}$ be the set of users.
For each user $u \in \mathcal{U}$, we define their historical context as $\mathcal{H}_u = \{(x_i, y_i)\}_{i=1}^n$.
Here, $x_i$ denotes the task context (\eg a specific instruction or prompt) and $y_i$ represents the ground-truth response authored by the user.
This history $\mathcal{H}_u$ serves as the primary evidence for inducing $u$'s implicit preferences.

In this work, we focus on the reference-free personalized evaluation setting, which mirrors real-world scenarios where a user's ground-truth response for a given task context may be unavailable.
Given a new task input $x$ and a model-generated response $\hat{y} = \mathcal{G}(x, \mathcal{H}_u)$, where $\mathcal{G}$ is a personalized generator (\eg an LLM), the objective is to assess how $\hat{y}$ align with the user's instrinsic preferences.
Unlike general evaluation, this task measures whether $\hat{y}$ aligns with the specific user preferences as established in their history $\mathcal{H}_u$.

Consequently, we define a scoring function $\Phi$ to compute an personalized alignment score $S$:
\begin{equation}
    S = \Phi(\hat{y}; x, \mathcal{H}_u).
\end{equation}
The score $S$ is normalized to $[0, 1]$, with higher values indicating stronger alignment with the user's preferences.

\subsection{Rubric-based Evaluation}

To enable a verifiable and granular assessment of personalized alignment, we introduce a rubric-based evaluation strategy that decomposes holistic user preferences into explicit, interpretable dimensions.
Specifically, the assessment of $\hat{y}$ is mediated by a personalized rubric set $\mathcal{R}_u = \{r_1, r_2, \dots, r_m\}$ for user $u$, where each rubric $r_j$ represents a distinct analytical dimension of the user's preferences, distilled from $\mathcal{H}_u$.
The final personalized alignment score $S$ is computed as the normalized average of these binary judgments across all $m$ dimensions for user $u$:
\begin{equation} 
S = \frac{1}{m} \sum_{j=1}^{m} \mathbb{I}(\hat{y} \vDash r_j),
\end{equation}
where $\vDash$ is a binary satisfaction operator.
The indicator function $\mathbb{I}(\cdot)$ returns $1$ when the response $\hat{y}$ satisfies rubric $r_j$, and $0$ otherwise.
By aggregating these binary outcomes, the resulting score $S \in [0, 1]$ provides a quantitative measure of alignment depth, where the binary nature of each $\mathbb{I}(\cdot)$ ensures that every dimension of the user’s evaluative rubric is explicitly and transparently verified.

\section{\ours~(\oursfull)}

In this section, we first provide an overview of our proposed \textbf{\ours} (\oursfull) framework, followed by detailed introductions of its primary components.

\begin{figure*}[!ht]
    \centering
    \includegraphics[width=0.96\linewidth]{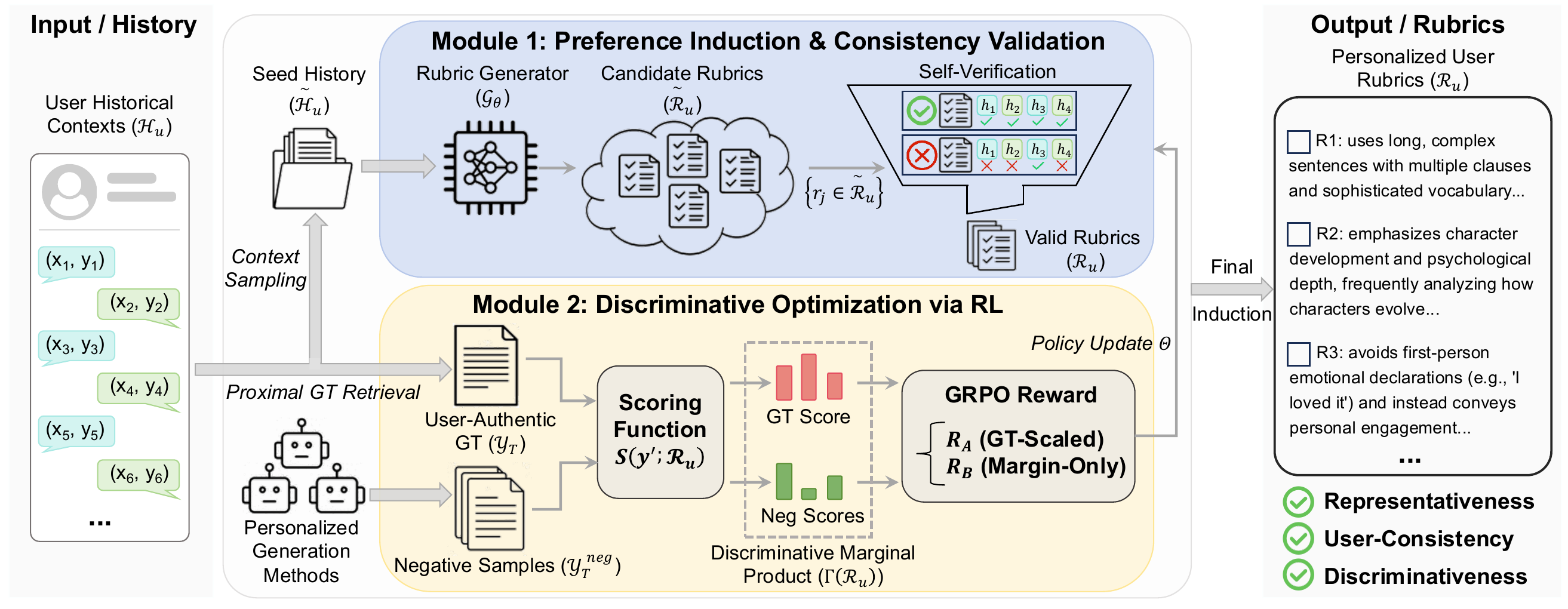}
    \vspace{-2mm}
    \caption{
    Overview of our proposed \ours framework for inducing personalized user rubrics for evaluating LLM personalization. \ours consists of two core modules: (1) Preference Induction \& Consistency Validation; (2) Discriminative Optimization via RL. The final induced rubrics serve as high-fidelity evaluation metrics for personalized text generation.
    }
    \label{main_method}
    \vspace{-2mm}
\end{figure*}

\subsection{Overview}

The \ours framework redefines personalized evaluation by shifting from an opaque, monolithic scoring paradigm to a structured, learnable, and verifiable process.
The primary objective of the framework is to transform unorganized, raw historical contexts $\mathcal{H}_u$ into user-specific rubrics $\mathcal{R}_u = \{r_1, r_2, \dots, r_m\}$ for personalized evaluation that rigorously satisfy three fundamental principles: \textit{Representativeness}, \textit{User-Consistency}, and \textit{Discriminativeness}.

As illustrated in Figure~\ref{main_method}, we implement a hierarchical pipeline consisting of two core modules:
\begin{itemize}[leftmargin=*, topsep=4pt, itemsep=0pt]
    \item \textbf{Preference Induction \& Consistency Validation.}
    This module focuses on the transition from raw historical contexts to a compact set of preference-aware rubrics for each user.
    The framework first performs a multi-dimensional analysis of a user's diverse historical contexts, capturing a comprehensive spectrum of multifaceted preferences that define their unique evaluative identity.
    The resulting candidate rubrics are subsequently validated through the user history, and only those that demonstrate persistent satisfaction across diverse historical contexts are retained.
    \item \textbf{Discriminative Optimization via RL.}
    To better disentangle user-specific stylistic signatures from historical noise, we incorporate a specialized learning stage that transforms rubric induction into a learnable capability via reinforcement learning (RL).
    By leveraging outputs from a representative, competitive set of personalized methods as negative samples, the framework optimizes a margin-based objective to maximize the scoring gap between authentic user-authored responses and competitive alternatives, thereby sharpening the decision boundaries of the induced rubrics.
\end{itemize}

Ultimately, this approach enables learning rubrics that precisely delineate user-aligned decision boundaries.
The resulting rubrics thus emerge as high-resolution, idiosyncratic evaluation metrics capable of capturing the subtle nuances of individual preferences.

\subsection{Preference Induction \& Consistency Validation}\label{preference_distillation}

To obtain highly representative rubrics, the framework must capture the multi-dimensional nature of user preferences embedded in $\mathcal{H}_u$.
We define a rubric generator $\mathcal{G}$ that induces a candidate set of rubrics $\tilde{\mathcal{R}}_u = \{r_1, r_2, \dots, r_k\}$ from a subset of the user history.
Unlike holistic summaries, each induced rubric $r_j$ is constrained to be atomic and single-dimensional, capturing a specific stylistic, structural, or semantic invariant.
Formally, we denote this process as:
\begin{equation}
    \tilde{\mathcal{R}}_u = \mathcal{G}(\tilde{\mathcal{H}}_u),
\end{equation}
where $\tilde{\mathcal{H}}_u \subset \mathcal{H}_u$ represents a diverse selection of user history as the seed history.
This decoupled structure facilitates modular validation, enabling the precise identification and filtering of user-inconsistent candidate rubrics.

To explicitly ensure that the generated rubrics reflect enduring user traits rather than transient or task-specific artifacts, we subject the candidate set $\tilde{\mathcal{R}}_u$ to a self-verification process.
This mechanism grounds rubrics in empirical evidence, reducing evaluative hallucinations.
Since a genuine user preference should remain stable across diverse histories, we evaluate each $r_j$ against every sample within the seed history $\tilde{\mathcal{H}}_u$. 
The final distilled rubric set $\mathcal{R}_u$ is defined as the subset of $\tilde{\mathcal{R}}_u$ that demonstrates universal satisfaction across the seed history:
\begin{equation}
    \mathcal{R}_u = \left\{ r_j \in \tilde{\mathcal{R}}_u \mid \frac{1}{|\tilde{\mathcal{H}}_u|} \sum_{(x_i, y_i) \in \tilde{\mathcal{H}}_u} \mathbb{I}(y_i \vDash r_j) \geq \tau \right\},
\end{equation}
where $\tau$ serves as the consistency threshold.
By setting a high $\tau$ (\textit{e.g.}, $1.0$), we enforce a strict requirement for temporal stability, ensuring that $r_j$ is not a reaction to a specific task but a persistent evaluative standard for user $u$.
This pruning process effectively filters out spurious patterns, yielding a stable and reliable evaluation benchmark that is consistent across diverse scenarios.

\subsection{Discriminative Optimization via RL}\label{discriminative_optimization}

While the distilled rubrics $\mathcal{R}_u$ capture the invariant core of user preferences, they still lack the resolution to disentangle idiosyncratic stylistic signatures from historical noise.
The goal of this stage is to instill a discriminative capacity within the induced rubrics, enabling them to effectively differentiate between the user’s authentic behavior and sophisticated AI imitations.
We formulate this as an RL problem where the generator $\mathcal{G}_{\theta}$ is optimized to maximize the scoring margin between these two distributions.
To this end, we introduce two variants of our framework, \ours-A and \ours-B, which differ in their reward formulation.

To ground the discriminative refinement, we pair the authentic response $y_T$ from the most proximal interaction $(x_T, y_T) \in \mathcal{H}_u \setminus \tilde{\mathcal{H}}_u$ with a set of negative samples $\mathcal{Y}_T^{neg} = \{y_{T,1}^{neg}, \dots, y_{T,m}^{neg}\}$ generated by representative methods.
Unlike the binary satisfaction logic used for consistency filtering in Section~\ref{preference_distillation}, we define a scoring function $S(y'; \mathcal{R}_u)$ that provides a soft probabilistic measure of how well a response $y'$ aligns with the generated rubrics.
The total score is computed as the mean alignment across the valid rubric set: $S(y'; \mathcal{R}_u) = \frac{1}{|\mathcal{R}_u|} \sum_{r \in \mathcal{R}_u} S(y'; r)$.
Utilizing this scoring paradigm, we further define the \textit{Discriminative Margin Product} $\Gamma(\mathcal{R}_u)$ as the joint probability that the GT response is superior over all competitive negatives:
\begin{equation}
    \Gamma(\mathcal{R}_u) = \prod_{y' \in \mathcal{Y}_T^{neg}} \sigma\left( \kappa \cdot (S(y_T; \mathcal{R}_u) - S(y'; \mathcal{R}_u)) \right),
\end{equation}
where $\sigma(\cdot)$ is the sigmoid function and $\kappa$ is a scaling factor that sharpens the discriminative boundary.

\vspace{+5pt}
\noindent
\textbf{Reward Formulation: \ours-A vs. \ours-B.}

We investigate two variants of reward computation, differing in whether the reward incorporates the absolute satisfaction of user-authored ground-truth response:
\begin{itemize}[leftmargin=*, topsep=2pt, itemsep=0pt]
    \item \ours-A (GT-Scaled): $R_{A} = \lambda \cdot \mathcal{C} \cdot S(y_T; \mathcal{R}_u) \cdot \Gamma(\mathcal{R}_u)$. This variant integrates the relative margin with absolute alignment quality, favoring rubrics that are both strongly satisfied by the ground-truth and highly discriminative.
    \item \ours-B (Margin-Only): $R_{B} = \lambda \cdot \mathcal{C} \cdot \Gamma(\mathcal{R}_u)$. This variant isolates the discriminative signal by relying solely on the relative margin.
\end{itemize}
In both variants, $\mathcal{C}$ serves as a cardinality factor that normalizes the reward according to the number of valid rubrics, and $\lambda$ is a global scaling constant to control reward magnitude.

Overall, the choice between \ours-A and \ours-B represents a trade-off: \ours-A balances discriminativeness with absolute preference fidelity, while \ours-B prioritizes pure contrastive sensitivity.
In practice, both variants offer distinct utility, with \ours-A providing a robust, grounded benchmark for stable alignment and \ours-B enabling the high-resolution recovery of the most subtle, idiosyncratic signatures that define a user’s unique evaluative identity.

\vspace{+5pt}
\noindent
\textbf{Policy Optimization.}

To optimize the rubric generator $\mathcal{G}_\theta$, we employ Group Relative Policy Optimization (GRPO), which allows the model to explore the complex space of evaluative criteria without the overhead of a separate value function.
For each $\tilde{\mathcal{H}}_u$, we sample $K$ rubric sets and maximize the following objective:
\begin{equation}
\small
\begin{gathered}
    \mathcal{J}(\theta) = \mathbb{E} \Bigg[ \frac{1}{K} \sum_{i=1}^K \min \bigg( \frac{\mathcal{G}_\theta (\mathcal{R}_u^i | \tilde{\mathcal{H}}_u)}{\mathcal{G}_{\theta_{\text{old}}} (\mathcal{R}_u^i | \tilde{\mathcal{H}}_u)} A_i, \\ \text{clip}\left(\frac{\mathcal{G}_\theta (\mathcal{R}_u^i | \tilde{\mathcal{H}}_u)}{\mathcal{G}_{\theta_{\text{old}}} (\mathcal{R}_u^i | \tilde{\mathcal{H}}_u)}, 1-\epsilon, 1+\epsilon\right) A_i \bigg) \Bigg] - \beta \mathbb{D}_{\text{KL}}(\mathcal{G}_\theta || \mathcal{G}_{\text{ref}}),
\end{gathered}
\end{equation} 
where $\epsilon$ and $\beta$ are hyperparameters governing the clipping range and the strength of the KL regularization, respectively. 
The $\text{clip}(\cdot)$ operator enforces a trust-region to ensure stable policy updates, while the KL term $\mathbb{D}_{\text{KL}}$ regularizes the generator toward the reference policy $\mathcal{G}_{\text{ref}}$. 
The relative advantage $A_k$ is computed by normalizing the reward $R(\mathcal{R}_u^k)$ within the sampled group:
$A_i = (R_i - \bar{R}) / \sigma_R$, 
where $R(\cdot) \in \{R_A, R_B\}$ is evaluated against the most proximal interaction $(x_T, y_T)$.
Here $\bar{R}$ and $\sigma_R$ denote the mean and standard deviation of rewards within the sampled group, respectively.

\subsection{Rubric Induction}

During inference, the optimized rubric generator $\mathcal{G}_\theta$ induces preference-aware personalized rubrics for user $u$, conditioned on the user seed history $\tilde{\mathcal{H}}_u$.
To further ensure these rubrics reflect enduring user preferences, we formalize the induction as a joint generation-verification process.
The final rubric set $\mathcal{R}_u$ consists of rubrics that satisfy an empirical consistency threshold $\tau$ across the history:
\begin{equation}
    \mathcal{R}_u = \left\{ r_j \in \mathcal{G}_\theta(\tilde{\mathcal{H}}_u) \mid \frac{1}{|\tilde{\mathcal{H}}_u|} \sum_{(x_i, y_i) \in \tilde{\mathcal{H}}_u} \mathbb{I}(y_i \vDash r_j) \geq \tau \right\}.
\end{equation}
The resulting rubrics $\mathcal{R}_u$ satisfy the core requirements of \textit{representativeness}, \textit{user-consistency}, and \textit{discriminativeness}. 
This yields a reliable, high-fidelity evaluation framework that effectively distinguishes authentic personalized resonance from AI outputs in downstream generation tasks.

\section{Experiment}

\subsection{Experimental Setup}\label{exp_setup}

\par
\noindent
\textbf{Datasets.}
To benchmark the rubrics induced by \ours across diverse personalized scenarios, we conduct experiments on three representative text generation tasks: \textit{Amazon Review Generation}\footnote{\url{https://huggingface.co/datasets/SnowCharmQ/DPL-main}}~\citep{amazon2023}, \textit{Reddit Topic Writing}\footnote{\url{https://huggingface.co/datasets/LongLaMP/LongLaMP}}~\citep{reddit}, and \textit{News Headline Generation}\footnote{\url{https://lamp-benchmark.github.io/}}~\citep{news}.
These tasks were selected to provide a comprehensive testbed for assessing how different model-generated baselines align with rubrics induced by \ours, covering both short-form and long-form personalized text generation.
Additional details and statistical analyses of datasets are provided in Appendix~\ref{apd_dataset}.

\vspace{+6pt}
\par
\noindent
\textbf{Baselines.}
We benchmark the performance of \ours against a diverse and representative suite of personalized generation methods, including in-context learning (Non, RAG, Non-Think, RAG-Think), supervised fine-tuning (SFT), reinforcement learning (GRPO), and a hybrid approach (SFT+GRPO).
Additionally, we incorporate user-authored ground-truth (GT) as the gold standard here.
A broader range of baselines, along with their detailed descriptions, is provided in Appendix~\ref{apd_baseline}.

\vspace{0.5em}
\par
\noindent
\textbf{Backbone Models.}
In main experiments, we use \texttt{Qwen3-8B}~\citep{qwen3} and \texttt{Qwen3-235B-A22B-Instruct} as backbones for rubric induction. Specifically, \texttt{Qwen3-8B} is employed for both zero-shot rubric induction and as a learnable model for our proposed \ours framework.
For the downstream personalized generation tasks, we use \texttt{Qwen3-8B} as the primary generator across all methods to ensure a consistent and fair comparison, following prior work~\citep{tagpr}.
Additional experiments with more backbone models are provided in Appendix~\ref{apd_exp}.

\vspace{0.5em}
\par
\noindent
\textbf{Evaluation Details.}
To systematically analyze the quality of induced rubrics for users in personalized evaluation, we consider the following primary metrics:
\begin{itemize}[leftmargin=*, topsep=2pt, itemsep=0pt]
    \item \textbf{\textit{User-level Accuracy}}.
    We implement a user-level evaluation protocol by employing a log-probability-based scoring scheme, where a strong evaluator LLM (\texttt{Qwen3-235B-A22B-Instruct}) is employed to score each rubric dimension by softmax-normalized ``Yes/No'' token probabilities, considering a dimension satisfied if $p(\text{Yes}) > 0.5$.
    We use an all-or-nothing aggregation, reporting the mean pass rate over test samples, where a sample scores 1 only if all personalized rubrics are satisfied.
    \item \textbf{\textit{User Coverage}}.
    This metric quantifies the breadth of applicability of our framework.
    It is defined as the proportion of users for whom the framework successfully generates a non-empty, valid set of rubrics.
\end{itemize}
To further quantify the evaluative resolution of the induced rubrics, we report \textit{Max-Diff}, defined as the difference between the \textit{user-level accuracy} of GT and that of the strongest baseline.
For all \textit{user-level accuracy} results, we report the mean and standard deviation over five independent runs.
In addition, we perform paired t-tests to examine whether GT responses achieve significantly higher \textit{user-level accuracy} than competitive baselines.

\vspace{0.5em}
\par
\noindent
\textbf{Implementation Details.}
We benchmark five variants for personalized user rubric induction here:
\begin{itemize}[leftmargin=*, topsep=2pt, itemsep=0pt]
    \item \textbf{\textit{LM-8B}}: A pre-trained \texttt{Qwen3-8B} model serving as the rubric generator. 
    \item \textbf{\textit{LM-235B}}: A pre-trained \texttt{Qwen3-235B-A22B-Instruct} model serving as the rubric generator.
    \item \textbf{\textit{\ours-0}}: Optimized only using the scoring function $S(y_i; \mathcal{R}_u)$.
    \item \textbf{\textit{\ours-A}}: Optimized using the GT-scaled reward ($R_A$).
    \item \textbf{\textit{\ours-B}}: Optimized using the margin-only reward ($R_B$).
\end{itemize}
\vspace{+2pt}
Detailed hyperparameter configurations and implementation settings are described in Appendix~\ref{apd_impl}. 
The prompts employed for rubric generation and evaluation are presented in Appendix~\ref{ov_prompt}.

\newcommand{\meanstd}[2]{$\scalebox{0.9}{#1}_{\pm \scalebox{0.52}{#2}}$}
\newcommand{\tablesmall}{\fontsize{15}{18}\selectfont}

\definecolor{color0}{HTML}{d1d9e1}
\definecolor{colorA}{HTML}{f6f1f9}
\definecolor{colorB}{HTML}{eef2fd}

\definecolor{pos}{RGB}{35, 139, 69}
\definecolor{neg}{RGB}{153, 52, 4}

\begin{table*}[!t]
    \centering
    \caption{
    Evaluation results of induced rubrics across three personalized text generation tasks. GT denotes the \textit{user-level accuracy} of user-authentic responses, while \textit{Max-Diff} indicates the difference between GT and the strongest non-GT alternative. Within each rubric evaluation group, the highest result is bolded, and the second-best is underlined. The symbol * indicates p-value < 0.05 in t-tests.
    The \textit{user coverage} is reported as a percentage.
    }
    \vspace{-2mm}
    \label{main_results}
    \fontsize{16}{18}\selectfont
    \setlength{\tabcolsep}{1.2pt}
    
    \setlength{\aboverulesep}{0pt}
    \setlength{\belowrulesep}{0pt}
    \renewcommand{\arraystretch}{1.58} 
    
    \resizebox{\linewidth}{!}{
    \begin{tabular}{c c *{3}{c c >{\columncolor{color0!15}}c >{\columncolor{colorA!40}}c >{\columncolor{colorB!40}}c}}
        \toprule
        & & \multicolumn{5}{c}{\textit{Amazon Review Generation}} & \multicolumn{5}{c}{\textit{Reddit Topic Writing}} & \multicolumn{5}{c}{\textit{News Headline Generation}} \\
        \cmidrule(lr){3-7} \cmidrule(lr){8-12} \cmidrule(lr){13-17}
        
        & & \tablesmall \mbox{LM-8B} & \tablesmall \mbox{LM-235B} & \tablesmall \mbox{\ours-0} & \tablesmall \textbf{\mbox{\ours-A}} & \tablesmall \textbf{\mbox{\ours-B}} 
          & \tablesmall \mbox{LM-8B} & \tablesmall \mbox{LM-235B} & \tablesmall \mbox{\ours-0} & \tablesmall \textbf{\mbox{\ours-A}} & \tablesmall \textbf{\mbox{\ours-B}} 
          & \tablesmall \mbox{LM-8B} & \tablesmall \mbox{LM-235B} & \tablesmall \mbox{\ours-0} & \tablesmall \textbf{\mbox{\ours-A}} & \tablesmall \textbf{\mbox{\ours-B}} \\
        \midrule
        
        \multirow{9}{*}{\rotatebox{90}{\textbf{\textit{User-level Accuracy}}}} 
        & \textbf{GT} & \meanstd{\textbf{0.931}}{0.001} & \meanstd{\textbf{0.794}*}{0.003} & \meanstd{\textbf{1.000}}{0.000} & \meanstd{\textbf{0.931}*}{0.002} & \meanstd{\textbf{0.930}*}{0.001} & \meanstd{0.892}{0.000} & \meanstd{\textbf{0.840}}{0.003} & \meanstd{0.954}{0.001} & \meanstd{\textbf{0.848}*}{0.001} & \meanstd{\textbf{0.860}*}{0.002} & \meanstd{\underline{0.804}}{0.003} & \meanstd{0.765}{0.002} & \meanstd{\underline{0.998}}{0.000} & \meanstd{\textbf{0.832}*}{0.005} & \meanstd{\textbf{0.804}*}{0.002} \\
        \cmidrule(lr){2-17}
        & Non        & \meanstd{0.548}{0.002} & \meanstd{0.256}{0.002} & \meanstd{\textbf{1.000}}{0.000} & \meanstd{0.272}{0.002} & \meanstd{0.567}{0.001} & \meanstd{0.422}{0.000} & \meanstd{0.371}{0.001} & \meanstd{0.972}{0.000} & \meanstd{0.219}{0.000} & \meanstd{0.101}{0.001} & \meanstd{0.656}{0.001} & \meanstd{0.636}{0.002} & \meanstd{\underline{0.998}}{0.000} & \meanstd{0.331}{0.004} & \meanstd{0.414}{0.004} \\
        & RAG        & \meanstd{0.624}{0.002} & \meanstd{0.344}{0.002} & \meanstd{\underline{0.998}}{0.000} & \meanstd{0.325}{0.002} & \meanstd{0.616}{0.001} & \meanstd{0.615}{0.000} & \meanstd{0.531}{0.001} & \meanstd{\textbf{0.985}}{0.000} & \meanstd{0.267}{0.001} & \meanstd{0.234}{0.001} & \meanstd{0.740}{0.000} & \meanstd{0.714}{0.003} & \meanstd{\underline{0.998}}{0.000} & \meanstd{0.468}{0.004} & \meanstd{0.569}{0.005} \\
        & Non-Think  & \meanstd{0.601}{0.001} & \meanstd{0.280}{0.002} & \meanstd{\textbf{1.000}}{0.000} & \meanstd{0.225}{0.002} & \meanstd{0.579}{0.001} & \meanstd{0.506}{0.000} & \meanstd{0.408}{0.001} & \meanstd{0.505}{0.001} & \meanstd{0.196}{0.002} & \meanstd{0.184}{0.001} & \meanstd{0.681}{0.002} & \meanstd{0.636}{0.002} & \meanstd{0.996}{0.000} & \meanstd{0.253}{0.002} & \meanstd{0.408}{0.006} \\
        & RAG-Think  & \meanstd{0.740}{0.003} & \meanstd{0.420}{0.001} & \meanstd{\textbf{1.000}}{0.000} & \meanstd{0.330}{0.001} & \meanstd{0.669}{0.002} & \meanstd{0.707}{0.000} & \meanstd{0.617}{0.001} & \meanstd{0.818}{0.001} & \meanstd{0.296}{0.002} & \meanstd{0.300}{0.001} & \meanstd{0.705}{0.003} & \meanstd{0.718}{0.002} & \meanstd{\underline{0.998}}{0.000} & \meanstd{0.415}{0.003} & \meanstd{0.561}{0.005} \\
        & SFT        & \meanstd{0.827}{0.002} & \meanstd{0.642}{0.003} & \meanstd{\textbf{1.000}}{0.000} & \meanstd{0.877}{0.002} & \meanstd{0.880}{0.001} & \meanstd{0.803}{0.001} & \meanstd{0.764}{0.001} & \meanstd{0.971}{0.001} & \meanstd{0.760}{0.004} & \meanstd{0.790}{0.002} & \meanstd{\textbf{0.816}}{0.001} & \meanstd{\textbf{0.787}}{0.002} & \meanstd{\textbf{0.999}}{0.000} & \meanstd{\underline{0.814}}{0.004} & \meanstd{\underline{0.750}}{0.004} \\
        & GRPO       & \meanstd{\underline{0.930}}{0.001} & \meanstd{\underline{0.787}}{0.001} & \meanstd{\textbf{1.000}}{0.000} & \meanstd{0.832}{0.002} & \meanstd{0.918}{0.001} & \meanstd{0.891}{0.001} & \meanstd{0.798}{0.003} & \meanstd{\underline{0.982}}{0.000} & \meanstd{0.268}{0.001} & \meanstd{0.484}{0.001} & \meanstd{0.783}{0.002} & \meanstd{0.745}{0.003} & \meanstd{\underline{0.998}}{0.000} & \meanstd{0.496}{0.004} & \meanstd{0.591}{0.008} \\
        & SFT+GRPO   & \meanstd{0.917}{0.003} & \meanstd{0.759}{0.001} & \meanstd{\textbf{1.000}}{0.000} & \meanstd{\underline{0.905}}{0.003} & \meanstd{\underline{0.927}}{0.001} & \meanstd{0.899}{0.001} & \meanstd{\underline{0.836}}{0.001} & \meanstd{0.971}{0.000} & \meanstd{\underline{0.770}}{0.002} & \meanstd{\underline{0.824}}{0.000} & \meanstd{0.773}{0.001} & \meanstd{\underline{0.774}}{0.002} & \meanstd{\textbf{0.999}}{0.000} & \meanstd{0.775}{0.004} & \meanstd{0.742}{0.010} \\
        \cmidrule(lr){2-17}
        & \textit{Max-Diff} & \textcolor{pos}{$+0.001$} & \textcolor{pos}{$+0.007$} & \textcolor{neg}{$-0.000$} & \textcolor{pos}{$+0.026$} & \textcolor{pos}{$+0.003$} & \textcolor{neg}{$-0.007$} & \textcolor{pos}{$+0.004$} & \textcolor{neg}{$-0.031$} & \textcolor{pos}{$+0.078$} & \textcolor{pos}{$+0.036$} & \textcolor{neg}{$-0.012$} & \textcolor{neg}{$-0.022$} & \textcolor{neg}{$-0.001$} & \textcolor{pos}{$+0.018$} & \textcolor{pos}{$+0.054$} \\
        \midrule
        \multicolumn{2}{l}{\textbf{\textit{User Coverage}}} 
        & 49.1\% & 84.6\% & 100.0\% & 99.3\% & 99.5\% & 75.0\% & 81.5\% & 100.0\% & 99.9\% & 99.4\% & 66.9\% & 87.1\% & 100.0\% & 99.0\% & 100.0\% \\
        \bottomrule
    \end{tabular}
    }
\end{table*}

\subsection{Main Results}\label{main_exp}

We assess the quality of the generated personalized user rubrics by applying them as evaluation criteria for both GT and a suite of baseline outputs.
As illustrated in Table~\ref{main_results}, our analysis consists of a \textit{user-level accuracy} evaluation across three personalized text generation datasets alongside a detailed assessment of \textit{user coverage}, from which we draw the following observations:
\begin{itemize}[leftmargin=*, topsep=2pt, itemsep=0pt]
    \item 
    Relying on hand-crafted yet static prompts within pre-trained LLMs as rubric generators, including variants LM-8B and LM-235B, proves inadequate for personalized evaluation, as evidenced by GT \textit{user-level accuracy} that can occasionally be lower than generic baselines.
    This indicates a clear misalignment between the induced rubrics and users’ true preferences, underscoring the critical necessity of framing personalized evaluation as a learning objective for better alignment.
    \item 
    \ours-0, which excludes the Discriminative Margin Product, fails to provide sufficient evaluative resolution.
    Despite achieving high \textit{user-level accuracy} for GT and other personalized baselines, it possesses virtually no discriminative capability, often leading to overly permissive rubrics that cannot distinguish between authentic user responses and AI outputs.
    This further validates the indispensable role of the Discriminative Margin Product in enforcing high-fidelity alignment during the RL process.
    \item 
    Across all three personalized text generation tasks, our primary frameworks, \ours-A and \ours-B, achieve absolute scores in GT \textit{user-level accuracy}, and consistently outperform all compared baselines while maintaining robust discriminative power.
    These variants effectively establish a high margin of resolution between authentic user preferences and various competitive baselines.
    These results demonstrate that the induced rubrics are not only empirically grounded but also possess high diagnostic fidelity, ultimately highlighting their substantial utility for the rigorous evaluation of personalized generation tasks.
    \item 
    In terms of practical viability, both LM-8B and LM-235B, relying on hand-crafted, static prompts with pre-trained models, exhibit low \textit{user coverage} due to frequent generation of unusable or hallucinated criteria.
    In contrast, our learned variants maintain near-complete \textit{user coverage} of approximately 100\%.
    This demonstrates a significant improvement in reliability, enabling effective deployment across diverse users in real-world personalized text generation tasks.
\end{itemize}

\subsection{Ablation Studies}

To further examine the contribution of each core design in our framework, we conduct ablation studies on the Amazon Review Generation task using \ours-A as the base configuration.

\begin{wraptable}{r}{0.48\linewidth}
  \vspace{-1.2em}
  \centering
  \caption{
  Ablation studies of \ours-A on Amazon Review Generation.
  }
  \vspace{-0.6em}
  \label{tab_ablation}
  \scriptsize
  \renewcommand{\arraystretch}{1.08}
  \setlength{\tabcolsep}{4pt}
  \resizebox{0.96\linewidth}{!}{
  \begin{tabular}{@{}lccc@{}}
    \toprule
    \textbf{Method} & \textbf{GT} & \textbf{Max-Diff} $\uparrow$ & \textbf{User Coverage} $\uparrow$ \\
    \midrule
    w/o PI & 0.953 & -0.018 & 81.1\% \\
    w/o CV & 0.916* & 0.011 & 90.5\% \\
    w/o RL & 0.931 & 0.001 & 49.1\% \\
    \midrule
    \rowcolor{colorA}
    \textbf{\ours-A} & 0.931* & \textbf{0.026} & \textbf{99.3\%} \\
    \bottomrule
  \end{tabular}
  }
  \vspace{-1.0em}
\end{wraptable}

As shown in Table~\ref{tab_ablation}, removing any core component degrades the effectiveness of PARL.
Removing Preference Induction (w/o PI) yields a negative M\textit{ax-Diff} despite high GT \textit{user-accuracy}, indicating that the rubrics become insufficiently personalized and fail to distinguish authentic user responses from strong alternatives.
Removing Consistency Validation (w/o CV) reduces both \textit{Max-Diff} and \textit{user coverage}, showing the importance of filtering unstable or task-specific criteria.
Finally, removing RL optimization (w/o RL) leads to near-zero \textit{Max-Diff} and much lower \textit{user coverage}, indicating that the rubrics lack sufficient discriminative power without margin-based optimization.

Overall, these results demonstrate the complementary roles of the three modules: PI captures representative user-specific criteria, CV improves rubric reliability, and RL sharpens the decision boundary between authentic user responses and competitive alternatives.

\subsection{In-Depth Analysis}

In this section, we conduct additional experiments to further analyze the design and effectiveness of our framework.

\subsubsection{Evaluation with LLM-as-a-Judge}

To highlight the limitations of holistic scoring in personalized contexts, we employ a standard LLM-as-a-judge approach as a baseline here to compare with our \ours.
For a fair comparison, the LLM judge operates in a reference-free setting and uses the same raw user history used for rubric generation to analyze user preferences in context.
The LLM-as-a-judge directly outputs a score on a $1-5$ scale, which we normalize to $[0, 1]$ for consistency.
We also use the same \texttt{Qwen3-235B-A22B-Instruct} as the judger model, with the prompt presented in Appendix~\ref{ov_prompt}.
We report results on three personalized text generation tasks here.

As shown in Figure~\ref{llm_as_a_judge_data}, while the LLM-as-a-judge yields reasonable representativeness and discriminativeness on \textit{Amazon Review}, its evaluative efficacy proves inconsistent across other tasks.
On \textit{Topic Writing}, the LLM judge fails to prioritize personalized resonance, assigning scores to user-authentic ground-truth responses that are actually surpassed by generic baseline outputs.
Furthermore, on \textit{News Headline}, the scores exhibit no discernible evaluative pattern, offering little distinction between personalized and generic content.
These findings indicate that standard LLM-as-a-judge approaches, even when provided with user historical contexts, struggle to serve as reliable, reference-free evaluators for personalized generation.
This further underscores the need for our shift toward \textit{Personalized Evaluation as Learning} paradigm, where the evaluator can be dynamically optimized to capture idiosyncratic user preferences rather than relying on static, general-purpose inference.

\subsubsection{Cross-Dataset Generalization Analysis}\label{ood}

To assess the generalizability and cross-domain robustness of our framework, we perform an out-of-domain (OOD) evaluation by applying a rubric generator originally trained on \textit{Amazon Review} to a distinct \textit{CDs\_and\_Vinyl} OOD category, with dataset details provided in Appendix~\ref{apd_dataset}.
In this configuration, the rubric generator is tasked with inducing user-specific rubrics for users entirely unseen during the training phase by leveraging their unique historical contexts.

\begin{figure}[!t]
\centering
\begin{minipage}[t]{0.48\linewidth}
    \centering
    \includegraphics[width=\linewidth]{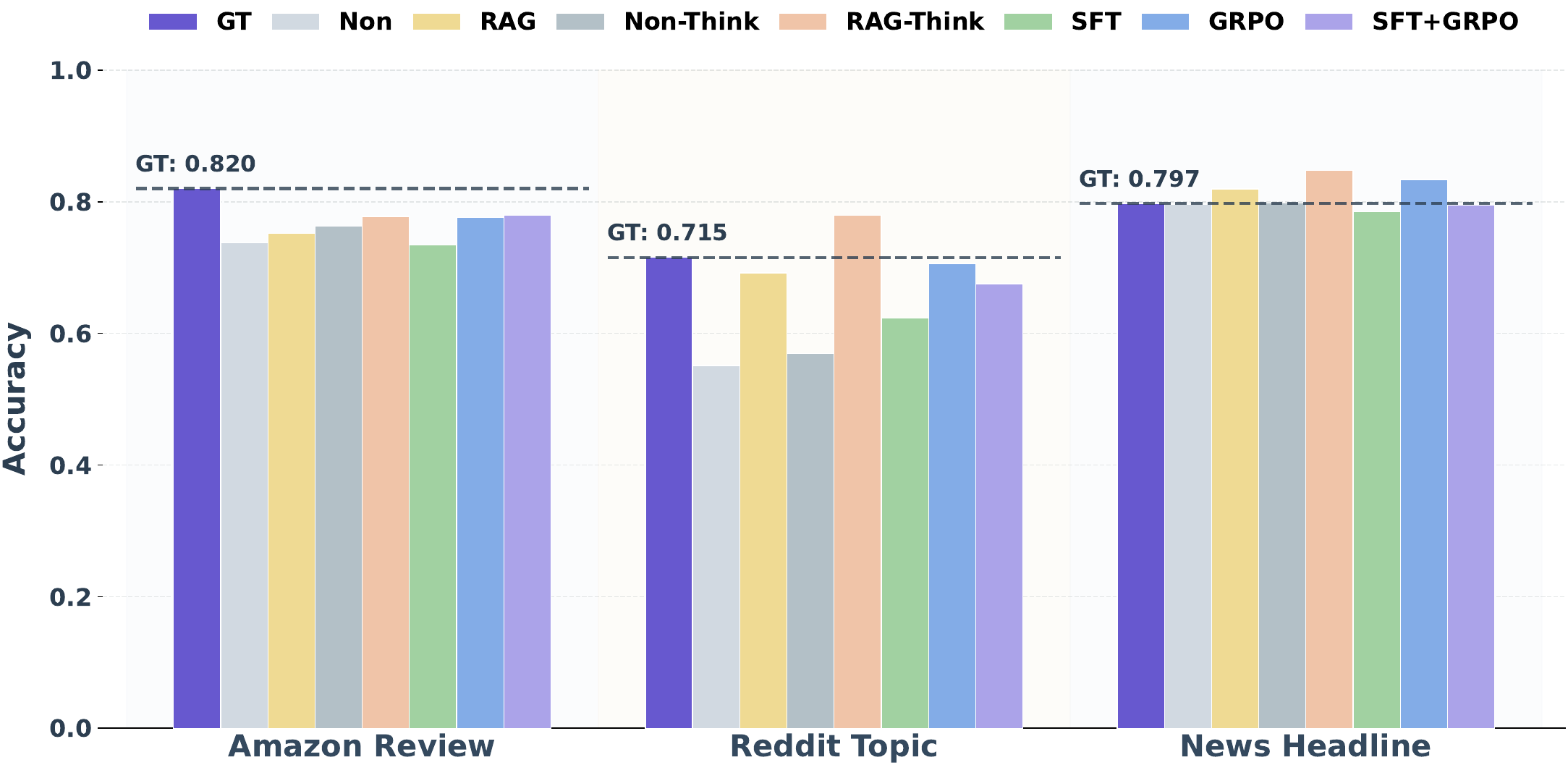}
    \vspace{-4mm}
    \caption{Results of LLM-as-a-judge evaluation scores across three datasets and various personalized generation methods.}
    \label{llm_as_a_judge_data}
\end{minipage}
\hfill
\begin{minipage}[t]{0.48\linewidth}
    \centering
    \includegraphics[width=\linewidth]{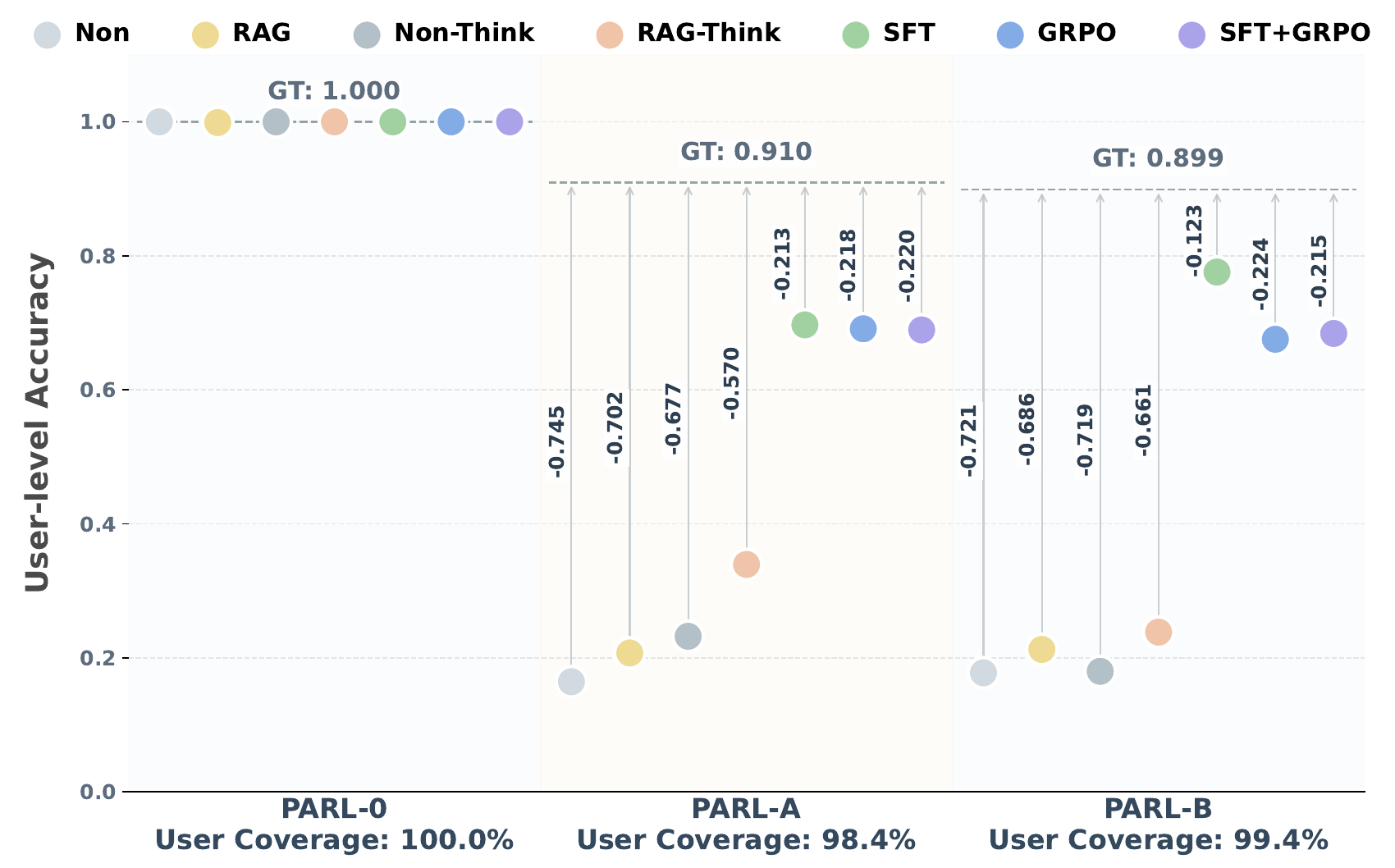}
    \vspace{-4mm}
    \caption{Evaluation of rubric induction variants on an out-of-domain dataset across various personalized generation methods.}
    \label{cds_data}
\end{minipage}
\end{figure}

We evaluate \ours-0, \ours-A, and \ours-B in this OOD setting to further analyze their performance.
As illustrated in Figure~\ref{cds_data}, despite achieving high \textit{user-level accuracy} and \textit{user coverage} on GT data, \ours-0 remains deficient in discriminative capability.
This further suggests that without the Discriminative Margin Product, the resulting rubrics are overly permissive and generic.
In contrast, both \ours-A and \ours-B consistently yield rubrics that demonstrate strong representativeness and discriminativeness simultaneously.
This strong performance indicates that the RL-driven induction process succeeds in distilling persistent stylistic invariants rather than memorizing dataset-specific surface patterns.
It further supports the effectiveness of our \textit{Personalized Evaluation as Learning} paradigm, demonstrating that the rubric generator can learn a transferable logic that captures the core of user identity.

\subsubsection{Intrinsic Analysis: Semantic Diversity and Specificity}

To further analyze the intrinsic quality and semantic patterns of the induced rubrics by our frameworks, we vectorize them using the \texttt{Qwen3-Embedding-8B}~\citep{qwen3embedding} model and project the resulting representations into a low-dimensional space.
This allows us to inspect their latent distribution, detect semantic collapse, and quantify evaluative specificity.

\begin{figure*}[t]
    \centering
    \includegraphics[width=0.88\linewidth]{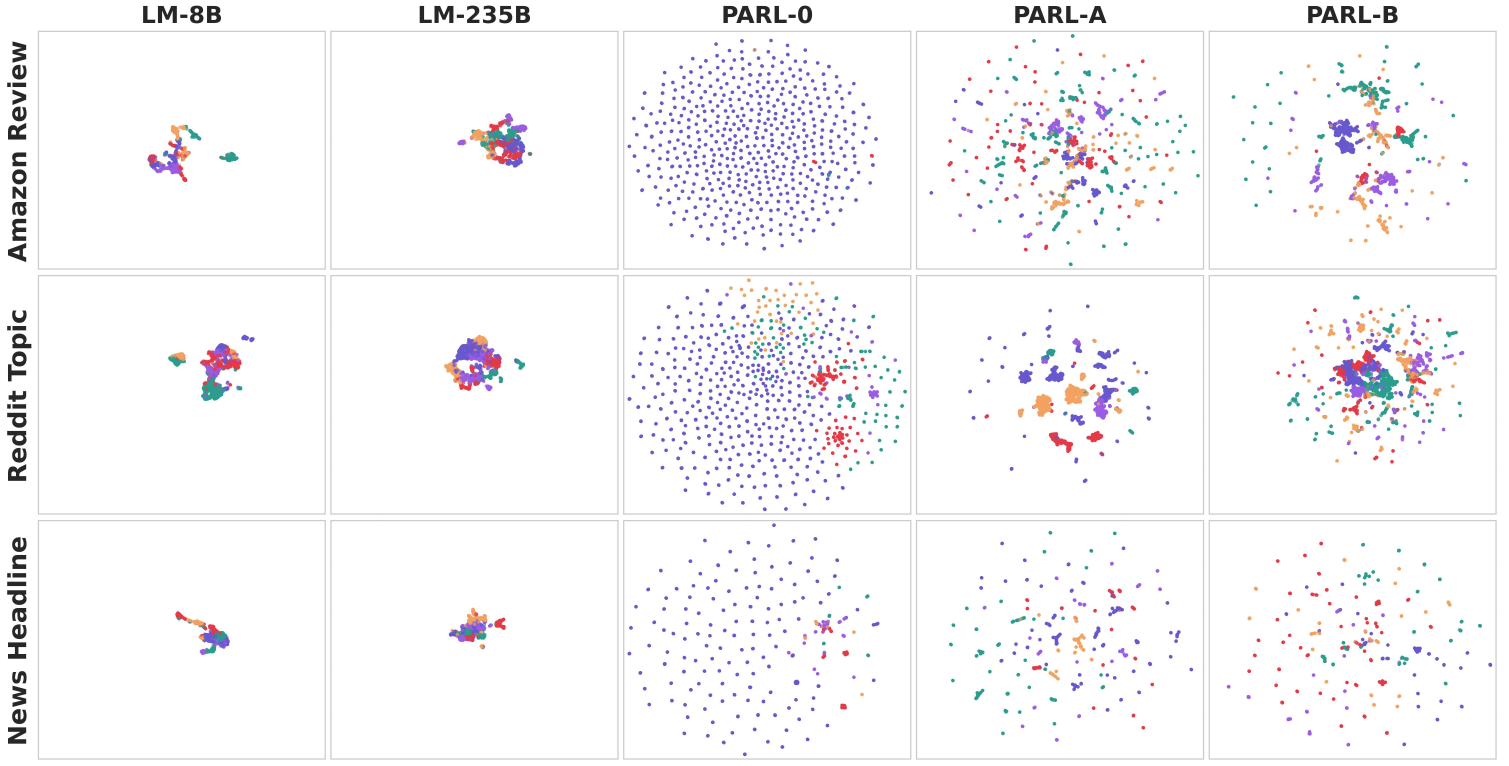}
    \vspace{-1mm}
    \caption{Intrinsic analysis of induced user rubrics across five rubric induction variants on three personalized text generation tasks. The visualization reveals a clear pattern shift from sparse, narrow clusters (LM-8B / LM-235B) and semantic collapse (dense redundancy in \ours-0) to well-distributed, structured scattering (tailored diversity in \ours-A and \ours-B).}
    \label{unsupervised_analysis}
    \vspace{-1mm}
\end{figure*}

As illustrated in Figure~\ref{unsupervised_analysis}, the rubric generators exhibit significant disparities in the semantic distributions of their induced rubrics.
Across all three personalized text generation tasks, for LM-8B and LM-235B, the generated rubrics are highly compressed within the semantic space, forming sparse and narrow clusters.
This reflects a limited capacity in pre-trained models to capture unique user traits, often resulting in rubrics restricted to a few common patterns.
\ours-0 demonstrates a clear semantic collapse; despite the high volume of rubrics, they form a dense, uniform sphere in the latent space.
This confirms that \ours-0 tends to produce redundant and generic high-quality criteria, leading to semantic convergence and a lack of discriminative resolution across users.
In contrast, \ours-A and \ours-B exhibit structured scattering, where rubrics are broadly distributed while forming clusters with distinct semantic features.
This indicates that our proposed framework successfully enforces the model to extract idiosyncratic stylistic traits and induce multi-dimensional, tailored, and personalized rubrics for diverse users, further establishing a high-fidelity alignment boundary to better facilitate the evaluation of personalized generation tasks.

\section{Discussion: Advantages over Existing Evaluation Paradigms}
Existing approaches to personalized evaluation span automatic metrics, human evaluation, and LLM-as-a-judge frameworks. Automatic metrics rely on surface-level matching and fixed references; human evaluation suffers from limited access to users’ true preferences; and LLM-as-a-judge approaches rely on static pre-trained models as judgers and heuristic prompts. As a result, these approaches lack interpretability, adaptability, or principled grounding in long-term user behavioral history.

In contrast, our \ours framework provides a reference-free and explicit evaluation framework that induces reusable, multi-dimensional rubrics directly grounded in user behavior. Unlike reward modeling methods, which entangle preference representation with policy optimization, PARL decouples evaluation from generation, enabling principled benchmarking across models, tasks, and training regimes. Moreover, by formulating rubric induction as a learning objective, PARL adapts dynamically to user-specific decision boundaries rather than relying on fixed or heuristic criteria, offering a robust and extensible foundation for personalized evaluation.

\section{Conclusion}
In this paper, we introduce the \textit{Personalized Evaluation as Learning} paradigm, which formulates personalized LLM evaluation as a learnable process grounded in three essential principles: \textit{Representativeness}, \textit{User-Consistency}, and \textit{Discriminativeness}. This paradigm addresses a key bottleneck in LLM personalization—the absence of adaptive and verifiable evaluation mechanisms for capturing nuanced individual preferences.
Under this paradigm, we propose \textbf{\ours}~(\oursfull), a learning-based framework that induces explicit, multi-dimensional evaluation rubrics from raw user histories, optimized with a discriminative reinforcement learning objective. \ours induces preference-aware rubrics that generalize across user historical contexts while preserving user-specific decision boundaries, enabling reliable identification of true personalized alignment.

Experiments on real-world personalized text generation tasks demonstrate that PARL consistently induces high-fidelity, reusable rubrics with strong user coverage and discriminative resolution. Further analysis shows that \ours generalizes well, and the induced rubrics are semantically diverse and user-specific.
Moving forward, this work advocates treating personalized evaluation as a learnable, adaptive process driven by the inference of preferences from user history rather than a static judgment, offering both academia and industry a novel and foundational paradigm for future research on personalized evaluation and alignment.

\bibliography{iclr2026_conference}
\bibliographystyle{iclr2026_conference}

\appendix

\appendix

\section{Limitations}



The effectiveness of our rubric generator depends heavily on the quality and quantity of available user behavioral history. In cold-start scenarios, where historical signals are sparse, the model may struggle to induce sufficiently detailed and stable criteria, constraining its performance despite robust results for active users. Moreover, our current framework assumes user preferences to be static signatures captured at a single point in time, without explicitly modeling their temporal evolution. Since human preferences and stylistic tendencies are inherently dynamic, future work could explore cross-user knowledge transfer, meta-learning strategies, or adaptive update mechanisms to enable rapid initialization for new users and continual refinement as preferences evolve.


Our investigation focuses on personalized text generation evaluation, while empirical exploration of personalized alignment and extensions to multi-modal settings remain open directions for future research.
More fundamentally, we acknowledge a philosophical constraint: only the individual user ultimately defines their own preferences, and our framework serves as a principled proxy for measurement rather than a perfect reflection of human subjectivity.
Although \ours achieves promising results, it has not yet attained a fully ideal state—where GT user-level accuracy reaches $1.0$ and user coverage remains consistently $100\%$.
This suggests that the nuanced preferences of inconsistent users remain difficult to capture, highlighting the need for future research to refine preference-aware strategies within this emerging paradigm.

\section{Ethical Considerations}

Our research involves the use of personalized user behavioral histories.
To protect user privacy, we use open-source datasets, ensuring full compliance with the original datasets' MIT license.
All data utilized in our experiments, including training sets for the rubric generator and the evaluation benchmarks, have been strictly de-identified.

While personalized evaluation aims to align with individual preferences, we acknowledge the risk that Large Language Models (LLMs) may inadvertently amplify social biases present in user data or pre-training corpora.
To mitigate this, our induced rubrics are designed to be transparent and interpretable, allowing for human auditing of the criteria.
We encourage users of our framework to implement additional safety filters to ensure that personalized criteria do not promote discriminatory or harmful content.
Besides, we strongly advocate for the responsible deployment of personalized evaluation frameworks.
The intended use of \ours is to improve the helpfulness and alignment of AI assistants, and we strictly prohibit its use for generating manipulative or malicious outputs.

\section{Dataset Details}\label{apd_dataset}

In this section, we provide more detailed descriptions and statistical analyses of our three representative and widespread evaluation tasks:

\begin{itemize}[leftmargin=*, itemsep=0pt]
    \item \textit{Amazon Review Generation}~\citep{amazon2023}:
    The objective of this task is to generate a detailed item review text conditioned on the input and the user profile. The input includes the item title and description, the user's rating for this item, and the title for this review. The user profile consists of the user's past reviews, including the corresponding item title and description, the rating for the item, the review title, and the review text.
    Specifically, we utilize the \textit{Movies\_and\_TV} and \textit{Books} categories for the main experiments in Section~\ref{main_exp}, while reserving \textit{CDs\_and\_Vinyl} as an out-of-domain (OOD) dataset for the analysis in Section~\ref{ood}.
    This task was processed by DPL~\citep{dpl}.
    \item \textit{Reddit Topic Writing}~\citep{reddit}:
    This task involves generating the full content of a Reddit post, based on the post's summary written by the author, and conditioned on a user profile consisting of the author's previous summary-content pairs.
    This task was processed by LongLaMP~\citep{longlamp}.
    \item \textit{News Headline Generation}~\citep{news}:
    This task requires the model to generate a headline for a given input news article, conditioned on a user profile that consists of the author's historical article-headline pairs.
    This task was processed by LaMP~\citep{lamp}.
\end{itemize}

\begin{table}[t]
\centering
\caption{Statistical data overview of three personalized text generation tasks.}
\vspace{-3mm}
\renewcommand{\arraystretch}{1.05}
\resizebox{0.7\textwidth}{!}{
\begin{tabular}{ccccc}
\hline
\textbf{Task} & \textbf{\# user} & \textbf{Dataset} & \textbf{\# data} & \textbf{len(data)} \\
\hline
\multirow{3}{*}{\makecell{Amazon\\Review}}
 & \multirow{3}{*}{2242}
 & \textit{train} & 16166 & 1379.65$\pm$1058.33 \\
 &  & \textit{val}   & 2242 & 1604.38$\pm$1529.67 \\
 &  & \textit{test}  & 2242 & 1632.54$\pm$1660.65 \\
\hline
\multirow{3}{*}{\makecell{Reddit\\Topic}}
 & \multirow{3}{*}{2452}
 & \textit{train} & 19337 & 1341.36$\pm$1126.73 \\
 &  & \textit{val}   & 2452 & 1886.86$\pm$1759.52 \\
 &  & \textit{test}  & 2452 & 1443.47$\pm$1286.57 \\
\hline
\multirow{3}{*}{\makecell{News\\Headline}}
 & \multirow{3}{*}{827}
 & \textit{train} & 5169 & 73.38$\pm$20.34 \\
 &  & \textit{val}   & 827 & 53.91$\pm$20.77 \\
 &  & \textit{test}  & 827 & 54.39$\pm$20.13 \\
\hline
\end{tabular}
}
\label{dataset_stat}
\vspace{-4mm}
\end{table}

In our experiments, we follow the temporal settings of these three datasets to configure the validation and test data splits.
Specifically, each test instance corresponds to a user’s most recent writing.
The detailed statistics of these datasets are summarized in Table~\ref{dataset_stat}.

\section{Baseline Details}\label{apd_baseline}

In this section, we provide more detailed descriptions of baselines benchmarked by \ours.
To comprehensively evaluate the effectiveness and robustness of our framework, we extend our analysis beyond methods featured in Section~\ref{exp_setup} to include a broader suite of competitive baselines.
These methods are categorized into four paradigms:
ICL (In-Context Learning), SFT (Supervised Fine-Tuning), RL (Reinforcement Learning), and Hybrid (SFT + RL).

\vspace{+2pt}
\noindent
\textbf{\textit{ICL:}}
\begin{itemize}[leftmargin=*, topsep=2pt, itemsep=0pt]
    \item \textbf{Non}: This method generates the target content without leveraging any user-specific information. The input to the model includes only the content-specific input.
    \item \textbf{RAG}~\citep{lamp}: This method enhances the personalized generation process by retrieving relevant historical user data from the user's profile, which is then formatted and included in the prompt context alongside the input.
    \item \textbf{PAG}~\citep{pag}: Building upon RAG, this method first summarizes the most recent historical instances from the user's profile into a compact personalized profile. The generated profile, along with the retrieved records, is included in the input to the LLM, allowing it to generate personalized content.
    \item \textbf{Non-Think}: This method enhances the basic Zero-Shot approach by enabling the thinking mode of LLM to analyze the input before generating the final content. No user history is leveraged.
    \item \textbf{RAG-Think}: This method enhances the RAG approach by generating an intermediate reasoning step that analyzes user preferences based on the retrieved history, and then leverages this reasoning to guide the personalized generation.
    \item \textbf{PAG-Think}: Building upon PAG, this method generates a reasoning path based on the summarized profile and retrieved records, guiding the subsequent generation of the personalized output.
    \item \textbf{R2P}~\citep{r2p}: This method enhances model reasoning for personalization by incorporating a Hierarchical Reasoning Thought Template, Reasoning Process Intervention, and Self-Referencing Module to guide the LLM toward structured and user-aligned outputs.
\end{itemize}

\vspace{+2pt}
\noindent
\textbf{\textit{SFT:}}
\begin{itemize}[leftmargin=*, topsep=2pt, itemsep=0pt]
    \item \textbf{SFT}~\citep{lamp}: This method directly trains the LLM on the full training dataset using Supervised Fine-Tuning (SFT), where the input prompt for each instance is augmented with the user history.
    \item \textbf{NextQuill}~\citep{nextquill}: This is an SFT-based method grounded in Causal Preference Modeling. It enhances personalization by aligning model-side causal preference effects with ground-truth data, ensuring the model focuses only on preference-driven tokens.
\end{itemize}

\newpage
\noindent
\textbf{\textit{RL:}}
\begin{itemize}[leftmargin=*, topsep=2pt, itemsep=0pt]
    \item \textbf{GRPO}~\citep{deepseekmath}: This method applies the Group Relative Policy Optimization (GRPO) RL framework to explicitly enhance the model's ability to generate effective personalization reasoning paths. All input prompts are augmented with the user's historical data. The reward used for GRPO training is defined as the sum of ROUGE-1 and METEOR scores.
    \item \textbf{GSPO}~\citep{gspo}: Similar to GRPO, this method employs the Group Sequence Policy Optimization (GSPO) RL framework. The input data is the same as GRPO. The reward used for GSPO training is defined as the sum of ROUGE-1 and METEOR scores.
    \item \textbf{PrLM}~\citep{prlm}: This is an RL-based method that trains the LLM for personalized RAG to explicitly reason over retrieved user profiles. It uses a contrastively trained personalization reward model to guide GRPO learning.
\end{itemize}

\vspace{+6pt}
\noindent
\textbf{\textit{Hybrid:}}
\begin{itemize}[leftmargin=*, topsep=2pt, itemsep=0pt]
    \item \textbf{TagPR}~\citep{tagpr}: A framework that enhances personalization reasoning by ``tagging the thought''. It employs a synergistic SFT-RL training strategy, guided by a composite reward signal that integrates tag-based structural constraints and a Personalization Reward Model with User Embeddings.
    \item \textbf{SFT+GRPO}: This method combines SFT and GRPO in a two-stage pipeline, first capturing general user styles via SFT and then refining the model using GRPO to better align with user preferences. The reward used during the RL stage is defined as the sum of ROUGE-1 and METEOR scores.
\end{itemize}

\section{Evaluation Details}\label{apd_evaluation}

In addition to the metrics described in Section~\ref{exp_setup}, we introduce a more granular measure to evaluate the fine-grained alignment of the generated rubrics:

\begin{itemize}[leftmargin=*, topsep=2pt, itemsep=0pt]
\item \textbf{\textit{Rubric-level Accuracy}}.
This metric provides a high-resolution assessment of rubric satisfaction by evaluating individual dimensions independently rather than using all-or-nothing aggregation.
It is calculated as the mean satisfaction rate across the entire pool of generated rubrics, where each specific rubric is scored using the same log-probability-based scheme ($p(\text{Yes}) > 0.5$) to determine if the response fulfills that particular stylistic constraint.
\end{itemize}

\section{Implementation Details}\label{apd_impl}

We provide more details about our implementation in this section.

\vspace{+2pt}
\noindent
\textbf{\textit{Rubric Induction:}}

To balance historical representativeness with inference efficiency, the rubric generator is conditioned on up to 10 of the latest interactions from $\mathcal{H}_u$.
For the RL stage, optimize the policy using GRPO with a group size $K=5$, a clipping threshold $\epsilon=0.2$, and a KL-penalty coefficient $\beta=0.01$.
In the reward function, the scaling factor $\kappa$ is set to $20$ to sharpen the discriminative margin, the reward scaling constant $\lambda$ is set to $10$, and the reward is modulated by a cardinality factor $\mathcal{C} = \min(1, |\mathcal{R}_u|/2)$.

We train the rubric generation model using the \texttt{verl}~\citep{verl} framework with a GRPO-based~\citep{deepseekmath} optimization objective.
Training is conducted with a batch size of $8$, a maximum prompt length of $10240$ tokens, and a maximum response length of $2048$ tokens.
Optimization is carried out using Adam~\citep{adam} with a learning rate of $1\times10^{-5}$ and a cosine learning rate scheduler.
For rollout generation, we employ \texttt{vLLM}~\citep{vllm} with nucleus sampling ($p=0.95$), temperature $0.6$, and top-$k$ sampling ($k=20$), producing $5$ samples per prompt.
For rubric generation, we adopt the same \texttt{vLLM}-based deployment configuration.
The selected baselines for computing \textit{Discriminative Margin Product} include: \textit{Non}, \textit{RAG}, \textit{SFT}, \textit{GRPO}, \textit{SFT+GRPO}.

\newpage
\noindent
\textbf{\textit{Personalized Text Generation:}}

For downstream personalized text generation implementation, we note that several baselines do not have official open-source implementations and are not built upon the same backbone model~\citep{r2p,tagpr}.
We therefore reproduce these methods and ensure that all baselines are evaluated under unified training and inference settings to guarantee a fair comparison.
For baselines that require user history, we adopt a BM25~\citep{bm25} retriever and consistently retrieve the top-4 user history items.

For SFT-based methods, we set the learning rate to $5\times10^{-6}$ and the weight decay to $0.025$, with a cosine learning rate scheduler and a warmup ratio of $0.01$.
For RL-based methods, we set the learning rate to $1\times10^{-6}$ with a cosine learning rate scheduler, and we sample 5 responses for each prompt during the policy rollout stage.
For hybrid methods, we use the same settings as the SFT- and RL-based methods in the respective stages.

Training is performed on 8 H800 GPUs, while generation is conducted using vLLM on a single H800 GPU with nucleus sampling ($p=0.95$), temperature $0.6$, and top-$k$ sampling ($k=20$).

\section{Additional Experiments}\label{apd_exp}

\subsection{Average Number of Rubrics per User}

We report the average number of rubrics induced per user across various generator variants, with the results summarized in Table~\ref{user_avg_num}.
The data indicate that standard pre-trained models generally yield a limited number of valid rubrics.
This constrained output highlights their relatively weak capacity to capture representative user signatures from raw historical contexts.

While \ours-0 generates a significantly higher volume of rubrics, these tend to be overly general and lack the discriminative resolution necessary for high-fidelity evaluation.
In contrast, \ours-A and \ours-B maintain a more moderate and reasonable rubric count.
This balance suggests that our learning-based process successfully filters out redundant or generic noise to distill a concise yet potent set of user-specific preferences.

\begin{table}[!t]
\centering
\caption{Average number of valid rubrics induced per user across three datasets.}
\vspace{-3mm}
\label{user_avg_num}
\setlength{\tabcolsep}{2pt}
\renewcommand{\arraystretch}{1.1}
\resizebox{0.72\linewidth}{!}{
\begin{tabular}{@{}l ccccc >{\columncolor{color0!15}}c >{\columncolor{colorA!40}}c >{\columncolor{colorB!40}}c@{}}
\toprule
Dataset & \makecell{LM-\\8B} & \makecell{LM-\\235B} & \makecell{LM-\\30B-IT} & \makecell{LM-30B\\-Think} & \makecell{GPT-4o\\-mini} & \makecell{\ours\\-0} & \makecell{\ours\\-A} & \makecell{\ours\\-B} \\ \midrule
Amazon Review & 0.71 & 1.81 & 0.44 & 2.27 & 2.11 & 10.00 & 4.01 & 3.40 \\
Reddit Post   & 1.20 & 1.46 & 0.66 & 2.65 & 2.64 & 9.26  & 4.55 & 4.60 \\
News Headline & 1.04 & 2.02 & 1.01 & 6.04 & 2.40 & 9.99  & 4.98 & 5.34 \\ \bottomrule
\end{tabular}
}
\vspace{-5mm}
\end{table}
\newcommand{\meanstdbig}[2]{$\scalebox{0.98}{#1}_{\pm \scalebox{0.7}{#2}}$}

\begin{wraptable}{r}{0.50\linewidth}
  \vspace{-1.2em}
  \centering
  \caption{
  Comparison of different cardinality constraints during rubric generator training.
  }
  \vspace{-0.6em}
  \label{cardinality}
  \scriptsize
  \setlength{\tabcolsep}{4pt}
  \renewcommand{\arraystretch}{1.02}
  \resizebox{0.96\linewidth}{!}{
  \begin{tabular}{@{}lcc@{}}
    \toprule
    \textbf{Paradigm} & \textbf{\textit{Threshold}} & \textbf{\textit{Linear}} \\
    \midrule
    \textbf{GT} & \meanstdbig{\textbf{0.9313}*}{0.0020} & \meanstdbig{\underline{0.9349}}{0.0010} \\
    \midrule
    Non       & \meanstdbig{0.2718}{0.0018} & \meanstdbig{0.3543}{0.0015} \\
    RAG       & \meanstdbig{0.3252}{0.0024} & \meanstdbig{0.3689}{0.0032} \\
    Non-Think & \meanstdbig{0.2250}{0.0017} & \meanstdbig{0.2956}{0.0013} \\
    RAG-Think & \meanstdbig{0.3299}{0.0013} & \meanstdbig{0.3447}{0.0017} \\
    SFT       & \meanstdbig{0.8773}{0.0017} & \meanstdbig{\textbf{0.9414}}{0.0012} \\
    GRPO      & \meanstdbig{0.8322}{0.0018} & \meanstdbig{0.8318}{0.0009} \\
    SFT+GRPO  & \meanstdbig{\underline{0.9053}}{0.0027} & \meanstdbig{0.9140}{0.0007} \\
    \midrule
    \textit{Max-Diff} & \textcolor{pos}{$+0.0260$} & \textcolor{neg}{$-0.0065$} \\
    \textit{User Coverage} & 99.33\% & 99.96\% \\
    \textit{Avg. Num. / User} & 4.01 & 6.56 \\
    \bottomrule
  \end{tabular}
  }
  \vspace{-1.0em}
\end{wraptable}

\subsection{Cardinality Analysis}

To analyze the impact of the cardinality constraint $\mathcal{C}$ during training, we compare our primary threshold-based implementation on the \textit{Amazon Review} task, defined as $\mathcal{C} = \min(1, |\mathcal{R}_u|/2)$, against a linear variant where $\mathcal{C} = |\mathcal{R}_u|/10$.
This linear implementation is designed to incentivize the rubric generator to produce the maximum possible number of valid rubrics for evaluation.
Here, we utilize the reward function from \ours-A, and we report the \textit{User-level Accuracy}, \textit{User Coverage}, \textit{Max-Diff}, and the average number of valid rubrics induced per user as \textit{Avg. Num. / User}.

As shown in Table \ref{cardinality}, while the Linear paradigm significantly increases the average number of valid rubrics induced per user, it fails to yield the highest GT \textit{user-level accuracy}.
This indicates that simply maximizing the number of rubrics during training is suboptimal, suggesting that an excessive quantity of rubrics may introduce redundancy or generic criteria, diluting evaluative precision.
Instead, focusing on a reasonable volume of high-utility, discriminative rubrics is essential for effective personalized evaluation.

\subsection{Effect of History Number}

\begin{wraptable}{r}{0.46\linewidth}
  \vspace{-1.0em}
  \centering
  \caption{Effect of history number of \ours-A on Amazon Review Generation.}
  \vspace{-0.6em}
  \label{history_number}
  \scriptsize
  \renewcommand{\arraystretch}{1.08}
  \setlength{\tabcolsep}{4pt}
  \resizebox{\linewidth}{!}{
  \begin{tabular}{@{}lccc@{}}
    \toprule
    \textbf{Method} & \textbf{GT} & \textbf{Max-Diff} $\uparrow$ & \textbf{Coverage} $\uparrow$ \\
    \midrule
    \#His=0 & 0.953 & -0.018 & 81.1\% \\
    \#His=2 & 0.930 & -0.007 & 91.3\% \\
    \#His=5 & 0.918 & 0.002 & 96.7\% \\
    \#His=8 & 0.921* & 0.019 & 99.1\% \\
    \midrule
    \rowcolor{colorA}
    \textbf{\ours-A} & 0.931* & \textbf{0.026} & \textbf{99.3\%} \\
    \bottomrule
  \end{tabular}
  }
  \vspace{-1.0em}
\end{wraptable}

We further analyze how the number of historical contexts affects rubric induction on the Amazon Review Generation task using \ours-A as the base configuration. 
As shown in Table~\ref{history_number}, using no historical context results in a significantly negative \textit{Max-Diff}, confirming that user history is essential for inducing personalized rubrics. 
With only 2 historical contexts, the generator still fails to obtain a positive margin, suggesting that limited history provides insufficient evidence for stable preference extraction.
As the history size increases to 5 and 8, \textit{Max-Diff} and \textit{user coverage} improve consistently, with 8 histories approaching the full setting of 10 histories. 
These results indicate that \ours does not require extensive user histories, but a moderately sized seed history is important for inducing reliable and representative personalized rubrics.

\subsection{Experiments Results on User-level Accuracy}

In this section, we report results for all introduced baselines in Appendix~\ref{apd_baseline}.
Besides, we employ a broader set of pre-trained backbone models, including the following: \texttt{Qwen3-30B-A3B-Instruct} (\textbf{\textit{LM-30B-It}}), \texttt{Qwen3-30B-A3B-Thinking} (\textbf{\textit{LM-30B-Think}}), and \texttt{GPT-4o-mini}.
The complete experimental results on \textit{user-level accuracy} are shown in Table~\ref{user_level_acc}.

The results further confirm the overall superiority of our proposed \ours framework: authentic user-authored GT responses consistently achieve the highest \textit{user-level accuracy}, demonstrating statistical significance in alignment with true preferences.
Moreover, our framework establishes a clear discriminative margin across diverse baselines, effectively distinguishing high-fidelity personalization from AI outputs.
In contrast, existing pre-trained models fail to meet the required evaluative standards, often yielding inconsistent or overly permissive \textit{user-level accuracy}.
This gap further underscores the structural advantage of our learning-driven approach in capturing the nuances of LLM personalization.

\subsection{Experiments Results on Rubric-level Accuracy}

We provide the complete experimental results on \textit{rubric-level accuracy} as shown in Table~\ref{rubric_level_acc}.

Our framework, \ours, consistently ensures that authentic user-authored GT responses achieve the highest \textit{rubric-level accuracy} across all three personalized text generation tasks.
Beyond absolute accuracy, \ours demonstrates robust discriminative power, effectively establishing a clear evaluative margin between GT and competitive baselines.
These results confirm that the induced rubrics are both well-grounded and highly diagnostic for personalized assessment.

\subsection{Experiments Results on Other Metrics}

We further provide experimental results of benchmarked baselines on existing evaluation metrics: ROUGE-1~\citep{rouge}, ROUGE-L, BLEU\footnote{We use the standard open-source \texttt{SacreBLEU}~\citep{sacrebleu} library to calculate the BLEU score: \url{https://github.com/mjpost/sacrebleu}.}~\citep{bleu}, METEOR~\citep{meteor}, BERTScore\footnote{We adopt the \texttt{led-base-16384}~\citep{longformer} model to obtain embeddings.}~\citep{bertscore}, and EmbScore\footnote{We use two state-of-the-art embedding models (\texttt{Qwen3-Embedding-0.6B}~\citep{qwen3embedding} and \texttt{Qwen3-Embedding-4B}) to encode both user-authored ground-truth texts and generated outputs, compute the cosine similarity for each model, and report the average as the score here.}.

From the experimental results presented in Table~\ref{baseline_metrics}, we draw the following observations:
\begin{itemize}[leftmargin=*, topsep=2pt, itemsep=0pt]
    \item 
    ICL-based methods that utilize historical user data outperform non-personalized baselines across almost all evaluated metrics.
    Its performance gain is highly consistent with the trends observed in our proposed \ours metric, further validating that historical context is indispensable for aligning model outputs with individual user characteristics.
    \item 
    Both SFT and RL strategies effectively enhance the alignment between model outputs and actual user ground-truth responses.
    Notably, the hybrid SFT+RL paradigm demonstrates the most comprehensive improvements across the majority of metrics.
    \item 
    A more granular analysis reveals notable discrepancies in evaluation trends, where high performance in lexical-based metrics (e.g., ROUGE-1, ROUGE-L, and BLEU) does not consistently translate to superior results in semantic-based metrics (e.g., METEOR, BERTScore, and EmbScore). 
    These metrics often exhibit divergent trends across different methods, suggesting they capture distinct and potentially conflicting facets of text quality.
    This inconsistency prompts a necessary reassessment of traditional evaluation protocols, underscoring the limitations of these metrics in providing a holistic measure of personalized generation.
\end{itemize}

\section{Case Study}

\begin{figure}[t]
    \centering
    \includegraphics[width=1\linewidth]{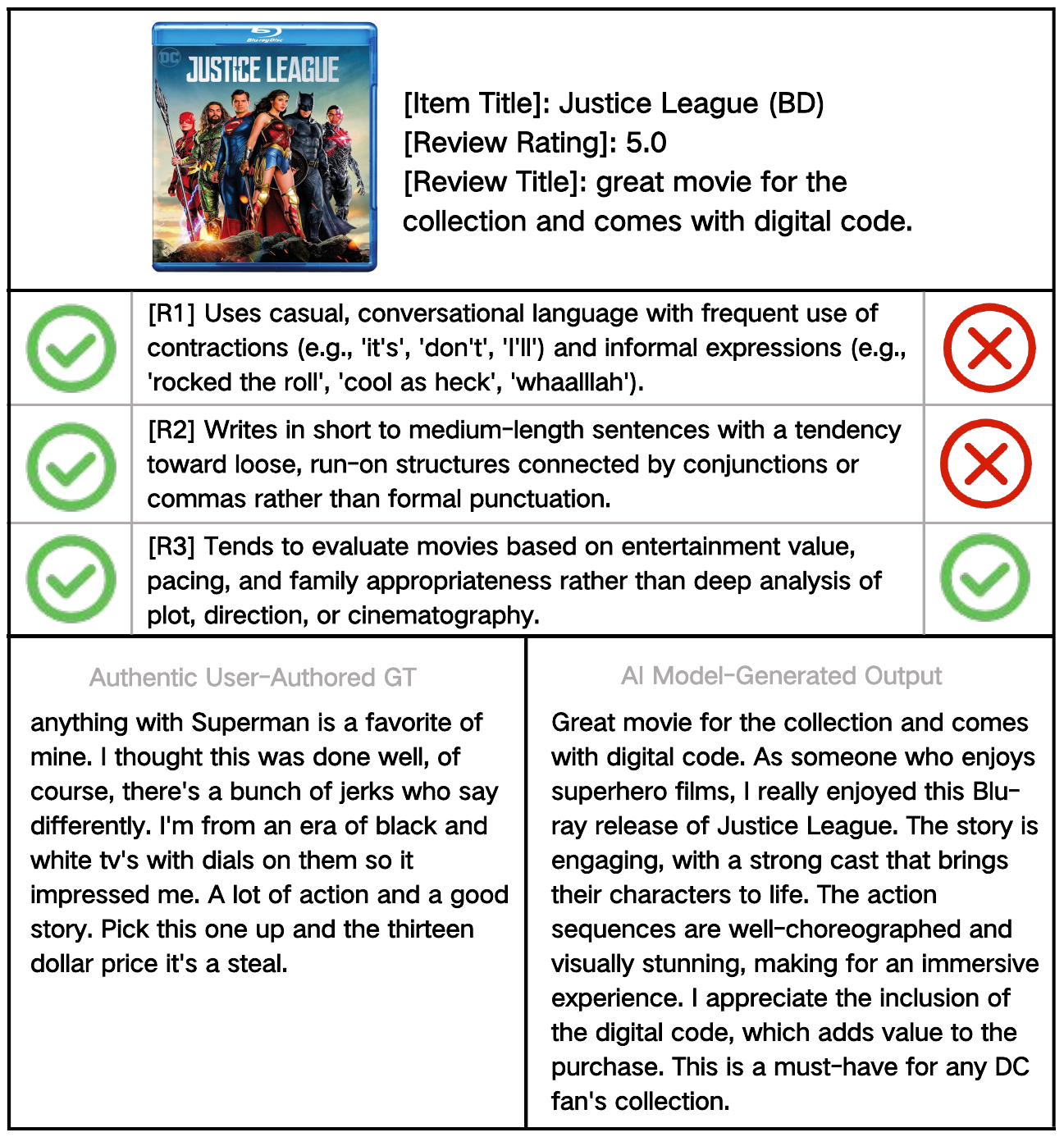}
    \vspace{-5mm}
    \caption{Qualitative analysis of induced rubrics on the \textit{Amazon Review} task. R1, R2, and R3 denote the induced rubrics for the target user. We evaluate whether the authentic user-authored GT and the AI model-generated output satisfy these rubrics.}
    \label{case_study}
    \vspace{-3mm}
\end{figure}

In this section, we present a case study from the \textit{Amazon Review} task to qualitatively illustrate how our induced rubrics evaluate personalized content, as shown in Figure~\ref{case_study}.
By comparing an authentic user-authored response (GT) with an AI-generated output, we demonstrate the framework's capacity to distinguish fine-grained stylistic nuances and personal preferences.
This analysis highlights the interpretability and diagnostic power of the rubrics in identifying the subtle misalignments often present in generic AI outputs.

\section{Overview of Prompts}\label{ov_prompt}

In this section, we illustrate the prompt design used in our framework \ours.

We present prompts used for generating personalized user rubrics in Table~\ref{prompt_generate_amazon}, Table~\ref{prompt_generate_reddit}, and Table~\ref{prompt_generate_news}.
To ensure the systematic and reliable generation of personalized user rubrics, our prompt design follows a structured hierarchical logic. 
Initially, the model is contextualized with its specific role and input data schema.
Subsequently, we define the core task under a rigorous set of drafting rules, encompassing atomicity, consistency, determinism, specificity, agnosticism, multi-dimensionality, and personalization.
Finally, the output is mandated in a structured JSON format to facilitate seamless downstream evaluation and programmatic parsing.

We present prompts used to evaluate the generated text and user-authored ground truth using personalized user rubrics generated by our framework in Table~\ref{prompt_evaluate_amazon}, Table~\ref{prompt_evaluate_topic}, and Table~\ref{prompt_evaluate_news}.

As shown in Table~\ref{prompt_llm_as_judge_amazon}, Table~\ref{prompt_llm_as_judge_topic}, and Table~\ref{prompt_llm_as_judge_news}, we also provide prompts used in the comparison LLM-as-a-judge experiments in Section~\ref{llm_as_judge} for reference.

\newcommand{\tablemid}{\fontsize{12}{12.5}\selectfont}

\begin{table*}[htbp]
    \centering
    \caption{Detailed evaluation results of induced rubrics across three personalized text generation tasks on \textit{user-level accuracy}.}
    \vspace{-3mm}
    \label{user_level_acc}
    \fontsize{11}{12.5}\selectfont
    \setlength{\tabcolsep}{3pt}
    
    \setlength{\aboverulesep}{0pt}
    \setlength{\belowrulesep}{0pt}
    \renewcommand{\arraystretch}{1.2} 
    
    \resizebox{1\linewidth}{!}{
    \begin{tabular}{c c l *{5}{c} >{\columncolor{color0!15}}c >{\columncolor{colorA!40}}c >{\columncolor{colorB!40}}c}
        \toprule
        \multicolumn{3}{c}{\textbf{Amazon Review}} & \tablemid LM-8B & \tablemid LM-235B & \tablemid LM-30B-IT & \tablemid LM-30B-Think & \tablemid GPT-4o-mini & \tablemid \ours-0 & \tablemid \textbf{\ours-A} & \tablemid \textbf{\ours-B} \\
        \midrule
        
        \multirow{18}{*}{\rotatebox{90}{\textbf{\textit{User-level Accuracy}}}} 
        & \multicolumn{2}{c}{\textbf{GT}} & \meanstd{\textbf{0.9307}}{0.0011} & \meanstd{\textbf{0.7943}*}{0.0030} & \meanstd{\textbf{0.9256}*}{0.0017} & \meanstd{0.9300}{0.0013} & \meanstd{0.9393}{0.0015} & \meanstd{\textbf{1.0000}}{0.0000} & \meanstd{0.9313}{0.0020} & \meanstd{\textbf{0.9295}*}{0.0006} \\
        \cmidrule(lr){2-11}
        
        & \multirow{7}{*}{ICL} 
        & Non               & \meanstd{0.5476}{0.0023} & \meanstd{0.2557}{0.0017} & \meanstd{0.4359}{0.0035} & \meanstd{0.6856}{0.0007} & \meanstd{0.6281}{0.0016} & \meanstd{\textbf{1.0000}}{0.0000} & \meanstd{0.2718}{0.0018} & \meanstd{0.5668}{0.0011} \\
        & & RAG               & \meanstd{0.6236}{0.0021} & \meanstd{0.3442}{0.0020} & \meanstd{0.5306}{0.0015} & \meanstd{0.7561}{0.0013} & \meanstd{0.6724}{0.0016} & \meanstd{0.9979}{0.0002} & \meanstd{0.3252}{0.0024} & \meanstd{0.6164}{0.0014} \\
        & & PAG               & \meanstd{0.6133}{0.0017} & \meanstd{0.3534}{0.0011} & \meanstd{0.5162}{0.0040} & \meanstd{0.7444}{0.0024} & \meanstd{0.6167}{0.0020} & \meanstd{0.9987}{0.0000} & \meanstd{0.2573}{0.0023} & \meanstd{0.5410}{0.0028} \\
        & & Non-Think         & \meanstd{0.6011}{0.0013} & \meanstd{0.2800}{0.0016} & \meanstd{0.4623}{0.0015} & \meanstd{0.6961}{0.0011} & \meanstd{0.6219}{0.0007} & \meanstd{\textbf{1.0000}}{0.0000} & \meanstd{0.2250}{0.0017} & \meanstd{0.5790}{0.0010} \\
        & & RAG-Think         & \meanstd{0.7396}{0.0029} & \meanstd{0.4203}{0.0011} & \meanstd{0.5942}{0.0030} & \meanstd{0.7892}{0.0006} & \meanstd{0.7333}{0.0012} & \meanstd{\underline{0.9997}}{0.0002} & \meanstd{0.3299}{0.0013} & \meanstd{0.6694}{0.0020} \\
        & & PAG-Think         & \meanstd{0.7100}{0.0026} & \meanstd{0.4266}{0.0010} & \meanstd{0.5628}{0.0010} & \meanstd{0.7875}{0.0012} & \meanstd{0.6368}{0.0019} & \meanstd{\textbf{1.0000}}{0.0000} & \meanstd{0.2397}{0.0048} & \meanstd{0.5614}{0.0021} \\
        & & R2P               & \meanstd{0.7640}{0.0016} & \meanstd{0.4473}{0.0007} & \meanstd{0.6633}{0.0018} & \meanstd{0.8138}{0.0008} & \meanstd{0.7552}{0.0016} & \meanstd{\textbf{1.0000}}{0.0000} & \meanstd{0.3538}{0.0016} & \meanstd{0.7100}{0.0019} \\
        \cmidrule(lr){2-11}
        
        & \multirow{2}{*}{SFT} 
        & SFT               & \meanstd{0.8269}{0.0023} & \meanstd{0.6416}{0.0033} & \meanstd{0.8124}{0.0015} & \meanstd{0.9057}{0.0016} & \meanstd{0.8671}{0.0018} & \meanstd{\textbf{1.0000}}{0.0000} & \meanstd{0.8773}{0.0017} & \meanstd{0.8800}{0.0012} \\
        & & NextQuill         & \meanstd{0.6733}{0.0015} & \meanstd{0.5449}{0.0014} & \meanstd{0.7124}{0.0019} & \meanstd{0.8116}{0.0027} & \meanstd{0.7316}{0.0019} & \meanstd{\textbf{1.0000}}{0.0000} & \meanstd{0.8673}{0.0011} & \meanstd{0.7880}{0.0011} \\
        \cmidrule(lr){2-11}
        
        & \multirow{3}{*}{RL} 
        & GRPO              & \meanstd{\underline{0.9304}}{0.0012} & \meanstd{\underline{0.7869}}{0.0007} & \meanstd{0.9182}{0.0030} & \meanstd{\textbf{0.9588}}{0.0017} & \meanstd{\textbf{0.9617}}{0.0009} & \meanstd{\textbf{1.0000}}{0.0000} & \meanstd{0.8322}{0.0018} & \meanstd{0.9182}{0.0010} \\
        & & GSPO              & \meanstd{0.9276}{0.0012} & \meanstd{0.7740}{0.0020} & \meanstd{\underline{0.9197}}{0.0017} & \meanstd{0.9516}{0.0015} & \meanstd{0.9475}{0.0018} & \meanstd{\textbf{1.0000}}{0.0000} & \meanstd{0.8006}{0.0021} & \meanstd{0.9191}{0.0011} \\
        & & PrLM              & \meanstd{0.9176}{0.0027} & \meanstd{0.7579}{0.0016} & \meanstd{0.8922}{0.0009} & \meanstd{0.9525}{0.0013} & \meanstd{0.9431}{0.0010} & \meanstd{\textbf{1.0000}}{0.0000} & \meanstd{0.7665}{0.0044} & \meanstd{0.8965}{0.0020} \\
        \cmidrule(lr){2-11}
        
        & \multirow{2}{*}{Hybrid} 
        & TagPR             & \meanstd{0.9265}{0.0013} & \meanstd{0.7625}{0.0007} & \meanstd{0.9063}{0.0029} & \meanstd{0.9482}{0.0011} & \meanstd{0.9424}{0.0020} & \meanstd{\textbf{1.0000}}{0.0000} & \meanstd{0.8658}{0.0031} & \meanstd{0.9118}{0.0009} \\
        & & SFT + GRPO         & \meanstd{0.9165}{0.0025} & \meanstd{0.7592}{0.0009} & \meanstd{0.9005}{0.0021} & \meanstd{\underline{0.9552}}{0.0019} & \meanstd{\underline{0.9512}}{0.0019} & \meanstd{\textbf{1.0000}}{0.0000} & \meanstd{\underline{0.9053}}{0.0027} & \meanstd{\underline{0.9272}}{0.0009} \\
        
        \midrule
        \multicolumn{3}{c}{\textbf{\textit{User Coverage}}} & 49.06\% & 84.61\% & 35.24\% & 78.64\% & 94.11\% & 100.00\% & 99.33\% & 99.85\% \\
        \bottomrule
        \\
        \toprule
        \multicolumn{3}{c}{\textbf{Reddit Topic}} & \tablemid LM-8B & \tablemid LM-235B & \tablemid LM-30B-IT & \tablemid LM-30B-Think & \tablemid GPT-4o-mini & \tablemid \ours-0 & \tablemid \textbf{\ours-A} & \tablemid \textbf{\ours-B} \\
        \midrule
        
        \multirow{18}{*}{\rotatebox{90}{\textbf{\textit{User-level Accuracy}}}} 
        & \multicolumn{2}{c}{\textbf{GT}} & \meanstd{0.8916}{0.0004} & \meanstd{\textbf{0.8395}}{0.0028} & \meanstd{0.8557}{0.0017} & \meanstd{0.8260}{0.0028} & \meanstd{0.8584}{0.0021} & \meanstd{0.9542}{0.0007} & \meanstd{\textbf{0.8480}*}{0.0014} & \meanstd{\textbf{0.8604}*}{0.0018} \\
        \cmidrule(lr){2-11}
        
        & \multirow{7}{*}{ICL} 
        & Non               & \meanstd{0.4218}{0.0002} & \meanstd{0.3714}{0.0007} & \meanstd{0.4201}{0.0023} & \meanstd{0.4175}{0.0014} & \meanstd{0.2846}{0.0011} & \meanstd{0.9724}{0.0002} & \meanstd{0.2190}{0.0003} & \meanstd{0.1013}{0.0006} \\
        & & RAG               & \meanstd{0.6150}{0.0003} & \meanstd{0.5307}{0.0012} & \meanstd{0.5275}{0.0015} & \meanstd{0.6241}{0.0015} & \meanstd{0.5406}{0.0009} & \meanstd{0.9849}{0.0004} & \meanstd{0.2666}{0.0007} & \meanstd{0.2340}{0.0013} \\
        & & PAG               & \meanstd{0.6404}{0.0006} & \meanstd{0.5817}{0.0020} & \meanstd{0.5185}{0.0009} & \meanstd{0.6719}{0.0021} & \meanstd{0.5520}{0.0024} & \meanstd{\underline{0.9852}}{0.0002} & \meanstd{0.1918}{0.0015} & \meanstd{0.1929}{0.0012} \\
        & & Non-Think         & \meanstd{0.5055}{0.0002} & \meanstd{0.4075}{0.0012} & \meanstd{0.4027}{0.0021} & \meanstd{0.4552}{0.0021} & \meanstd{0.3978}{0.0012} & \meanstd{0.5046}{0.0011} & \meanstd{0.1958}{0.0022} & \meanstd{0.1841}{0.0008} \\
        & & RAG-Think         & \meanstd{0.7070}{0.0004} & \meanstd{0.6170}{0.0012} & \meanstd{0.5799}{0.0028} & \meanstd{0.6806}{0.0007} & \meanstd{0.6463}{0.0020} & \meanstd{0.8184}{0.0006} & \meanstd{0.2965}{0.0015} & \meanstd{0.2998}{0.0013} \\
        & & PAG-Think         & \meanstd{0.6898}{0.0008} & \meanstd{0.6294}{0.0011} & \meanstd{0.5405}{0.0020} & \meanstd{0.6914}{0.0021} & \meanstd{0.6063}{0.0010} & \meanstd{0.7845}{0.0011} & \meanstd{0.2637}{0.0013} & \meanstd{0.3049}{0.0013} \\
        & & R2P               & \meanstd{0.7986}{0.0002} & \meanstd{0.6736}{0.0007} & \meanstd{0.7305}{0.0006} & \meanstd{0.6879}{0.0021} & \meanstd{0.7320}{0.0020} & \meanstd{0.9472}{0.0005} & \meanstd{0.5296}{0.0027} & \meanstd{0.4817}{0.0026} \\
        \cmidrule(lr){2-11}
        
        & \multirow{2}{*}{SFT} 
        & SFT               & \meanstd{0.8028}{0.0007} & \meanstd{0.7642}{0.0011} & \meanstd{0.8343}{0.0014} & \meanstd{0.7577}{0.0020} & \meanstd{0.7248}{0.0013} & \meanstd{0.9706}{0.0007} & \meanstd{0.7600}{0.0044} & \meanstd{0.7899}{0.0016} \\
        & & NextQuill         & \meanstd{0.6879}{0.0016} & \meanstd{0.6612}{0.0038} & \meanstd{0.7927}{0.0030} & \meanstd{0.6635}{0.0018} & \meanstd{0.5626}{0.0018} & \meanstd{0.9803}{0.0002} & \meanstd{0.7365}{0.0018} & \meanstd{0.7705}{0.0006} \\
        \cmidrule(lr){2-11}
        
        & \multirow{3}{*}{RL} 
        & GRPO              & \meanstd{0.8909}{0.0011} & \meanstd{0.7985}{0.0026} & \meanstd{\underline{0.8577}}{0.0011} & \meanstd{0.8602}{0.0009} & \meanstd{0.8727}{0.0011} & \meanstd{0.9821}{0.0004} & \meanstd{0.2678}{0.0011} & \meanstd{0.4841}{0.0009} \\
        & & GSPO              & \meanstd{0.8948}{0.0009} & \meanstd{0.8038}{0.0015} & \meanstd{0.8462}{0.0025} & \meanstd{0.8552}{0.0016} & \meanstd{\underline{0.8728}}{0.0013} & \meanstd{\textbf{0.9933}}{0.0002} & \meanstd{0.3533}{0.0014} & \meanstd{0.4739}{0.0007} \\
        & & PrLM              & \meanstd{0.7677}{0.0021} & \meanstd{0.6137}{0.0012} & \meanstd{0.7375}{0.0017} & \meanstd{0.7612}{0.0009} & \meanstd{0.6683}{0.0014} & \meanstd{0.9646}{0.0003} & \meanstd{0.2896}{0.0026} & \meanstd{0.4936}{0.0009} \\
        \cmidrule(lr){2-11}
        
        & \multirow{2}{*}{Hybrid} 
        & TagPR             & \meanstd{\textbf{0.9067}}{0.0012} & \meanstd{0.8006}{0.0015} & \meanstd{0.8343}{0.0025} & \meanstd{\textbf{0.8875}}{0.0009} & \meanstd{\textbf{0.8902}}{0.0008} & \meanstd{0.9851}{0.0002} & \meanstd{0.4192}{0.0018} & \meanstd{0.6071}{0.0010} \\
        & & SFT + GRPO         & \meanstd{\underline{0.8988}}{0.0009} & \meanstd{\underline{0.8359}}{0.0008} & \meanstd{\textbf{0.8757}}{0.0014} & \meanstd{\underline{0.8741}}{0.0015} & \meanstd{0.8700}{0.0006} & \meanstd{0.9711}{0.0004} & \meanstd{\underline{0.7705}}{0.0018} & \meanstd{\underline{0.8243}}{0.0004} \\
        
        \midrule
        \multicolumn{3}{c}{\textbf{\textit{User Coverage}}} & 74.96\% & 81.53\% & 51.43\% & 85.73\% & 98.12\% & 100.00\% & 99.88\% & 99.43\% \\
        \bottomrule
        \\
        \toprule
        \multicolumn{3}{c}{\textbf{News Headline}} & \tablemid LM-8B & \tablemid LM-235B & \tablemid LM-30B-IT & \tablemid LM-30B-Think & \tablemid GPT-4o-mini & \tablemid \ours-0 & \tablemid \textbf{\ours-A} & \tablemid \textbf{\ours-B} \\
        \midrule
        
        \multirow{18}{*}{\rotatebox{90}{\textbf{\textit{User-level Accuracy}}}} 
        & \multicolumn{2}{c}{\textbf{GT}} & \meanstd{0.8043}{0.0029} & \meanstd{0.7647}{0.0019} & \meanstd{0.8184}{0.0000} & \meanstd{0.7198}{0.0006} & \meanstd{0.7653}{0.0010} & \meanstd{\underline{0.9976}}{0.0000} & \meanstd{\textbf{0.8325}*}{0.0053} & \meanstd{\textbf{0.8039}*}{0.0022} \\
        \cmidrule(lr){2-11}
        
        & \multirow{7}{*}{ICL} 
        & Non               & \meanstd{0.6561}{0.0007} & \meanstd{0.6364}{0.0016} & \meanstd{0.6893}{0.0000} & \meanstd{0.5939}{0.0008} & \meanstd{0.6192}{0.0015} & \meanstd{\underline{0.9976}}{0.0000} & \meanstd{0.3311}{0.0037} & \meanstd{0.4135}{0.0040} \\
        & & RAG               & \meanstd{0.7396}{0.0000} & \meanstd{0.7136}{0.0027} & \meanstd{0.7987}{0.0000} & \meanstd{0.6712}{0.0008} & \meanstd{0.7058}{0.0006} & \meanstd{\underline{0.9976}}{0.0000} & \meanstd{0.4679}{0.0043} & \meanstd{0.5693}{0.0053} \\
        & & PAG               & \meanstd{0.7707}{0.0029} & \meanstd{0.7425}{0.0023} & \meanstd{0.8201}{0.0009} & \meanstd{0.5961}{0.0005} & \meanstd{0.7187}{0.0012} & \meanstd{0.9964}{0.0000} & \meanstd{0.5541}{0.0081} & \meanstd{0.6784}{0.0049} \\
        & & Non-Think         & \meanstd{0.6810}{0.0018} & \meanstd{0.6356}{0.0024} & \meanstd{0.6937}{0.0000} & \meanstd{0.5747}{0.0006} & \meanstd{0.6137}{0.0005} & \meanstd{0.9964}{0.0000} & \meanstd{0.2535}{0.0018} & \meanstd{0.4082}{0.0062} \\
        & & RAG-Think         & \meanstd{0.7052}{0.0032} & \meanstd{0.7183}{0.0022} & \meanstd{0.7681}{0.0000} & \meanstd{0.6012}{0.0000} & \meanstd{0.6772}{0.0000} & \meanstd{\underline{0.9976}}{0.0000} & \meanstd{0.4151}{0.0035} & \meanstd{0.5608}{0.0048} \\
        & & PAG-Think         & \meanstd{0.7649}{0.0011} & \meanstd{0.7197}{0.0042} & \meanstd{0.7681}{0.0000} & \meanstd{0.5870}{0.0015} & \meanstd{0.6663}{0.0010} & \meanstd{\underline{0.9976}}{0.0000} & \meanstd{0.5053}{0.0034} & \meanstd{0.6839}{0.0049} \\
        & & R2P               & \meanstd{0.7928}{0.0014} & \meanstd{0.7536}{0.0023} & \meanstd{0.8249}{0.0000} & \meanstd{0.7323}{0.0005} & \meanstd{0.7646}{0.0000} & \meanstd{0.9964}{0.0000} & \meanstd{0.4449}{0.0066} & \meanstd{0.6537}{0.0049} \\
        \cmidrule(lr){2-11}
        
        & \multirow{2}{*}{SFT} 
        & SFT               & \meanstd{\underline{0.8156}}{0.0011} & \meanstd{\textbf{0.7867}}{0.0021} & \meanstd{0.8048}{0.0009} & \meanstd{\textbf{0.7482}}{0.0005} & \meanstd{0.7463}{0.0013} & \meanstd{\textbf{0.9988}}{0.0000} & \meanstd{\underline{0.8139}}{0.0039} & \meanstd{\underline{0.7497}}{0.0040} \\
        & & NextQuill         & \meanstd{0.7935}{0.0027} & \meanstd{0.7550}{0.0029} & \meanstd{0.8206}{0.0000} & \meanstd{0.7318}{0.0010} & \meanstd{0.7395}{0.0005} & \meanstd{\textbf{0.9988}}{0.0000} & \meanstd{0.8034}{0.0020} & \meanstd{0.7378}{0.0055} \\
        \cmidrule(lr){2-11}
        
        & \multirow{3}{*}{RL} 
        & GRPO              & \meanstd{0.7834}{0.0018} & \meanstd{0.7450}{0.0026} & \meanstd{\underline{0.8337}}{0.0000} & \meanstd{0.6299}{0.0006} & \meanstd{0.7319}{0.0005} & \meanstd{\underline{0.9976}}{0.0000} & \meanstd{0.4960}{0.0039} & \meanstd{0.5906}{0.0077} \\
        & & GSPO              & \meanstd{0.8108}{0.0018} & \meanstd{0.7722}{0.0018} & \meanstd{\textbf{0.8446}}{0.0000} & \meanstd{\underline{0.7399}}{0.0008} & \meanstd{\underline{0.7790}}{0.0006} & \meanstd{0.9964}{0.0000} & \meanstd{0.5587}{0.0039} & \meanstd{0.6788}{0.0066} \\
        & & PrLM              & \meanstd{0.8130}{0.0022} & \meanstd{0.7767}{0.0031} & \meanstd{0.8293}{0.0000} & \meanstd{0.6989}{0.0013} & \meanstd{\textbf{0.7851}}{0.0009} & \meanstd{\underline{0.9976}}{0.0000} & \meanstd{0.5111}{0.0068} & \meanstd{0.6455}{0.0083} \\
        \cmidrule(lr){2-11}
        
        & \multirow{2}{*}{Hybrid} 
        & TagPR             & \meanstd{\textbf{0.8174}}{0.0034} & \meanstd{\underline{0.7792}}{0.0020} & \meanstd{0.8096}{0.0000} & \meanstd{0.6626}{0.0008} & \meanstd{0.7749}{0.0005} & \meanstd{0.9964}{0.0000} & \meanstd{0.4847}{0.0061} & \meanstd{0.5555}{0.0042} \\
        & & SFT + GRPO         & \meanstd{0.7732}{0.0009} & \meanstd{0.7742}{0.0023} & \meanstd{0.8140}{0.0000} & \meanstd{0.7394}{0.0013} & \meanstd{0.7661}{0.0009} & \meanstd{\textbf{0.9988}}{0.0000} & \meanstd{0.7753}{0.0035} & \meanstd{0.7420}{0.0100} \\
        
        \midrule
        \multicolumn{3}{c}{\textbf{\textit{User Coverage}}} & 66.87\% & 87.06\% & 55.26\% & 98.55\% & 95.53\% & 100.00\% & 99.03\% & 100.00\% \\
        \bottomrule
    \end{tabular}
    }
\end{table*}
\begin{table*}[htbp]
    \centering
    \caption{Detailed evaluation results of induced rubrics across three personalized text generation tasks on \textit{Rubric-level Accuracy}.}
    \vspace{-3mm}
    \label{rubric_level_acc}
    \fontsize{16}{18}\selectfont
    \setlength{\tabcolsep}{3pt}
    
    \setlength{\aboverulesep}{0pt}
    \setlength{\belowrulesep}{0pt}
    \renewcommand{\arraystretch}{1.2} 
    
    \resizebox{1\linewidth}{!}{
    \begin{tabular}{c c l *{5}{c} >{\columncolor{color0!15}}c >{\columncolor{colorA!40}}c >{\columncolor{colorB!40}}c}
        \toprule
        \multicolumn{3}{c}{\textbf{Amazon Review}} & \tablesmall LM-8B & \tablesmall LM-235B & \tablesmall LM-30B-IT & \tablesmall LM-30B-Think & \tablesmall GPT-4o-mini & \tablesmall \ours-0 & \tablesmall \textbf{\ours-A} & \tablesmall \textbf{\ours-B} \\
        \midrule
        
        \multirow{18}{*}{\rotatebox{90}{\textbf{\textit{Rubric-level Accuracy}}}} 
        & \multicolumn{2}{c}{\textbf{GT}} & \underline{0.9447} & \textbf{0.8825} & \underline{0.9341} & 0.9717 & 0.9680 & \textbf{1.0000} & \textbf{0.9766} & \textbf{0.9768} \\
        \cmidrule(lr){2-11}
        
        & \multirow{7}{*}{ICL} 
        & Non               & 0.5971 & 0.3927 & 0.4468 & 0.8473 & 0.7758 & \textbf{1.0000} & 0.5652 & 0.8013 \\
        & & RAG               & 0.6675 & 0.5011 & 0.5593 & 0.8722 & 0.7965 & \underline{0.9998} & 0.6100 & 0.8289 \\
        & & PAG               & 0.6776 & 0.5299 & 0.5532 & 0.8711 & 0.7806 & \underline{0.9998} & 0.5410 & 0.7961 \\
        & & Non-Think         & 0.6593 & 0.4422 & 0.4863 & 0.8555 & 0.7855 & \textbf{1.0000} & 0.4994 & 0.8086 \\
        & & RAG-Think         & 0.7957 & 0.6061 & 0.6353 & 0.9086 & 0.8609 & \textbf{1.0000} & 0.6055 & 0.8628 \\
        & & PAG-Think         & 0.7775 & 0.6085 & 0.6130 & 0.9076 & 0.8126 & \textbf{1.0000} & 0.5134 & 0.8115 \\
        & & R2P               & 0.8158 & 0.6373 & 0.6981 & 0.9184 & 0.8761 & \textbf{1.0000} & 0.6402 & 0.8838 \\
        \cmidrule(lr){2-11}
        
        & \multirow{2}{*}{SFT} 
        & SFT               & 0.8567 & 0.7687 & 0.8278 & 0.9617 & 0.9257 & \textbf{1.0000} & 0.9595 & 0.9533 \\
        & & NextQuill         & 0.6958 & 0.6632 & 0.7285 & 0.9137 & 0.8116 & \textbf{1.0000} & 0.9574 & 0.9115 \\
        \cmidrule(lr){2-11}
        
        & \multirow{3}{*}{RL} 
        & GRPO              & \textbf{0.9485} & \underline{0.8780} & 0.9331 & \textbf{0.9839} & \textbf{0.9826} & \textbf{1.0000} & 0.9423 & 0.9736 \\
        & & GSPO              & 0.9428 & 0.8719 & \textbf{0.9352} & \underline{0.9823} & 0.9727 & \textbf{1.0000} & 0.9301 & 0.9747 \\
        & & PrLM              & 0.9371 & 0.8667 & 0.9108 & 0.9819 & 0.9729 & \textbf{1.0000} & 0.9138 & 0.9652 \\
        \cmidrule(lr){2-11}
        
        & \multirow{2}{*}{Hybrid} 
        & TagPR             & 0.9390 & 0.8606 & 0.9189 & 0.9798 & 0.9699 & \textbf{1.0000} & 0.9580 & 0.9728 \\
        & & SFT + GRPO         & 0.9397 & 0.8620 & 0.9149 & 0.9809 & \underline{0.9763} & \textbf{1.0000} & \underline{0.9687} & \underline{0.9752} \\

        \bottomrule
        \\
        \toprule
        \multicolumn{3}{c}{\textbf{Reddit Topic}} & \tablesmall LM-8B & \tablesmall LM-235B & \tablesmall LM-30B-IT & \tablesmall LM-30B-Think & \tablesmall GPT-4o-mini & \tablesmall \ours-0 & \tablesmall \textbf{\ours-A} & \tablesmall \textbf{\ours-B} \\
        \midrule
        
        \multirow{18}{*}{\rotatebox{90}{\textbf{\textit{Rubric-level Accuracy}}}} 
        & \multicolumn{2}{c}{\textbf{GT}} & 0.9180 & \textbf{0.8942} & \underline{0.8806} & 0.9313 & 0.9324 & 0.9933 & \textbf{0.9543} & \textbf{0.9559} \\
        \cmidrule(lr){2-11}
        
        & \multirow{7}{*}{ICL} 
        & Non               & 0.4501 & 0.4743 & 0.4265 & 0.6828 & 0.4975 & 0.9965 & 0.6024 & 0.3508 \\
        & & RAG               & 0.6546 & 0.6427 & 0.5557 & 0.8175 & 0.7373 & 0.9980 & 0.6198 & 0.4193 \\
        & & PAG               & 0.6871 & 0.6721 & 0.5471 & 0.8476 & 0.7589 & \textbf{0.9984} & 0.5471 & 0.3567 \\
        & & Non-Think         & 0.5273 & 0.5006 & 0.4166 & 0.7094 & 0.6042 & 0.9374 & 0.5578 & 0.3292 \\
        & & RAG-Think         & 0.7501 & 0.7012 & 0.5840 & 0.8455 & 0.8090 & 0.9773 & 0.6184 & 0.4418 \\
        & & PAG-Think         & 0.7392 & 0.7107 & 0.5557 & 0.8496 & 0.8007 & 0.9720 & 0.5827 & 0.4298 \\
        & & R2P               & 0.8293 & 0.7557 & 0.7422 & 0.8576 & 0.8595 & 0.9935 & 0.8058 & 0.6748 \\
        \cmidrule(lr){2-11}
        
        & \multirow{2}{*}{SFT} 
        & SFT               & 0.8507 & 0.8458 & 0.8566 & 0.9035 & 0.8575 & 0.9957 & 0.9306 & 0.9154 \\
        & & NextQuill         & 0.7491 & 0.7557 & 0.8160 & 0.8561 & 0.7481 & 0.9973 & 0.9257 & 0.9141 \\
        \cmidrule(lr){2-11}
        
        & \multirow{3}{*}{RL} 
        & GRPO              & 0.9231 & 0.8657 & 0.8769 & 0.9458 & \underline{0.9418} & 0.9977 & 0.7178 & 0.6699 \\
        & & GSPO              & 0.9204 & 0.8696 & 0.8609 & 0.9427 & 0.9411 & 0.9993 & 0.7848 & 0.7358 \\
        & & PrLM              & 0.8320 & 0.7409 & 0.7735 & 0.9032 & 0.8362 & 0.9959 & 0.7381 & 0.6301 \\
        \cmidrule(lr){2-11}
        
        & \multirow{2}{*}{Hybrid} 
        & TagPR             & \textbf{0.9292} & 0.8702 & 0.8517 & \textbf{0.9553} & \textbf{0.9488} & \underline{0.9982} & 0.7684 & 0.7997 \\
        & & SFT + GRPO         & \underline{0.9238} & \underline{0.8920} & \textbf{0.8948} & \underline{0.9525} & 0.9400 & 0.9959 & \underline{0.9356} & \underline{0.9396} \\

        \bottomrule
        \\
        \toprule
        \multicolumn{3}{c}{\textbf{News Headline}} & \tablesmall LM-8B & \tablesmall LM-235B & \tablesmall LM-30B-IT & \tablesmall LM-30B-Think & \tablesmall GPT-4o-mini & \tablesmall \ours-0 & \tablesmall \textbf{\ours-A} & \tablesmall \textbf{\ours-B} \\
        \midrule
        
        \multirow{18}{*}{\rotatebox{90}{\textbf{\textit{Rubric-level Accuracy}}}} 
        & \multicolumn{2}{c}{\textbf{GT}} & 0.8345 & 0.8679 & 0.8725 & 0.9397 & 0.8821 & \underline{0.9998} & \textbf{0.9461} & \textbf{0.9520} \\
        \cmidrule(lr){2-11}
        
        & \multirow{7}{*}{ICL} 
        & Non               & 0.6678 & 0.7525 & 0.7592 & 0.9029 & 0.7748 & \underline{0.9998} & 0.6510 & 0.7824 \\
        & & RAG               & 0.7743 & 0.8279 & 0.8629 & 0.9257 & 0.8383 & \underline{0.9998} & 0.7341 & 0.8652 \\
        & & PAG               & 0.8299 & 0.8697 & 0.8856 & 0.9105 & 0.8645 & 0.9996 & 0.8049 & 0.9120 \\
        & & Non-Think         & 0.6875 & 0.7519 & 0.7580 & 0.8907 & 0.7758 & 0.9996 & 0.5555 & 0.7765 \\
        & & RAG-Think         & 0.7569 & 0.8314 & 0.8367 & 0.9007 & 0.8343 & \underline{0.9998} & 0.6945 & 0.8742 \\
        & & PAG-Think         & 0.8206 & 0.8422 & 0.8427 & 0.9091 & 0.8348 & \underline{0.9998} & 0.7665 &  0.9154 \\
        & & R2P               & 0.8264 & 0.8542 & 0.8820 & 0.9445 & 0.8715 & 0.9996 & 0.7181 & 0.9023 \\
        \cmidrule(lr){2-11}
        
        & \multirow{2}{*}{SFT} 
        & SFT               & 0.8542 & \textbf{0.8840} & 0.8915 & \textbf{0.9487} & 0.8640 & \textbf{0.9999} & \underline{0.9256} & 0.9294 \\
        & & NextQuill         & 0.8345 & 0.8703 & 0.8951 & \underline{0.9469} & 0.8730 & \textbf{0.9999} & \underline{0.9256} & \underline{0.9362} \\
        \cmidrule(lr){2-11}
        
        & \multirow{3}{*}{RL} 
        & GRPO              & 0.8426 & 0.8631 & \underline{0.8987} & 0.9187 & 0.8660 & \underline{0.9998} & 0.7575 & 0.8821 \\
        & & GSPO              & \textbf{0.8657} & 0.8781 & \textbf{0.9023} & 0.9453 & 0.8806 & 0.9996 & 0.7861 & 0.9138 \\
        & & PrLM              & 0.8553 & 0.8781 & 0.8939 & 0.9379 & \underline{0.8892} & \underline{0.9998} & 0.7580 & 0.8995 \\
        \cmidrule(lr){2-11}
        
        & \multirow{2}{*}{Hybrid} 
        & TagPR             & \underline{0.8611} & \underline{0.8804} & 0.8808 & 0.9247 & \textbf{0.8917} & 0.9996 & 0.7142 & 0.8568 \\
        & & SFT + GRPO         & 0.8206 & 0.8769 & 0.8808 & 0.9457 & 0.8710 & \textbf{0.9999} & 0.9152 & 0.9357 \\

        \bottomrule
    \end{tabular}
    }
\end{table*}

\begin{table*}[t]
\centering
\small
\setlength{\tabcolsep}{0pt}
\caption{Experimental results of measured methods on six other metrics: ROUGE-1 (R-1), ROUGE-L (R-L), BLEU (BL.), METEOR (ME.), BERTScore (BS.), and EmbScore (EmbS.).}
\vspace{-3mm}
\label{baseline_metrics}

\begin{tabular*}{\textwidth}{@{\extracolsep{\fill}} ll ccc ccc ccc @{}}
\toprule
\multirow{2}{*}{\textbf{Category}} & \multirow{2}{*}{\textbf{Method}} & \multicolumn{3}{c}{\textbf{Amazon Review}} & \multicolumn{3}{c}{\textbf{Reddit Topic}} & \multicolumn{3}{c}{\textbf{News Headline}} \\
\cmidrule(lr){3-5} \cmidrule(lr){6-8} \cmidrule(lr){9-11}
& & R-1 & R-L & BL. & R-1 & R-L & BL. & R-1 & R-L & BL. \\
\midrule
\multirow{7}{*}{ICL} & Non & 0.2381 & 0.1305 & 1.2345 & 0.1951 & 0.1044 & 0.6381 & 0.1413 & 0.1247 & 3.0168 \\
& RAG & 0.2731 & 0.1426 & 2.0540 & 0.2671 & 0.1298 & 1.8512 & 0.1563 & 0.1413 & 4.1086 \\
& PAG & 0.2723 & 0.1400 & 2.0441 & 0.2737 & 0.1267 & 1.4741 & 0.1441 & 0.1280 & 3.5684 \\
& Non-Think & 0.2227 & 0.1220 & 1.0618 & 0.1928 & 0.1003 & 0.6875 & 0.1367 & 0.1186 & 2.7902 \\
& RAG-Think & 0.2579 & 0.1358 & 1.7980 & 0.2339 & 0.1122 & 1.0177 & 0.1361 & 0.1205 & 3.2062 \\
& PAG-Think & 0.2534 & 0.1319 & 1.5694 & 0.2405 & 0.1126 & 0.9469 & 0.1398 & 0.1237 & 3.0538 \\
& R2P & 0.2734 & 0.1459 & 2.6768 & 0.2375 & 0.1189 & 1.1350 & 0.1630 & 0.1493 & 4.6484 \\
\midrule
\multirow{2}{*}{SFT} & SFT & 0.3473 & 0.1724 & 7.1323 & 0.3257 & 0.1667 & 4.9135 & 0.2023 & 0.1869 & 6.4891 \\
& NextQuill & 0.3241 & \textbf{0.1831} & \textbf{9.1712} & 0.2918 & 0.1658 & 4.7346 & \textbf{0.2150} & \textbf{0.2005} & 6.8295 \\
\midrule
\multirow{3}{*}{RL} & GRPO & 0.3735 & 0.1740 & 8.6424 & 0.3429 & 0.1610 & 4.0917 & 0.1877 & 0.1686 & 5.2834 \\
& GSPO & 0.3700 & 0.1722 & 8.4999 & 0.3443 & 0.1485 & 2.9119 & 0.1863 & 0.1714 & 5.6436 \\
& PrLM & 0.3654 & 0.1688 & 8.1936 & 0.3256 & 0.1486 & 2.9488 & 0.1914 & 0.1714 & 5.5231 \\
\midrule
\multirow{2}{*}{Hybrid} & TagPR & 0.3760 & 0.1724 & 8.6316 & 0.3424 & 0.1536 & 3.7206 & 0.1917 & 0.1735 & 5.7818 \\
& SFT+GRPO & \textbf{0.3829} & 0.1775 & 8.9420 & \textbf{0.3648} & \textbf{0.1739} & \textbf{5.2343} & 0.2120 & 0.1974 & \textbf{6.9968} \\
\bottomrule
\end{tabular*}

\vspace{1.2em} 

\begin{tabular*}{\textwidth}{@{\extracolsep{\fill}} ll ccc ccc ccc @{}}
\toprule
\multirow{2}{*}{\textbf{Category}} & \multirow{2}{*}{\textbf{Method}} & \multicolumn{3}{c}{\textbf{Amazon Review}} & \multicolumn{3}{c}{\textbf{Reddit Topic}} & \multicolumn{3}{c}{\textbf{News Headline}} \\
\cmidrule(lr){3-5} \cmidrule(lr){6-8} \cmidrule(lr){9-11}
& & ME. & BS. & EmbS. & ME. & BS. & EmbS. & ME. & BS. & EmbS. \\
\midrule
\multirow{7}{*}{ICL} & Non & 0.1361 & 0.4717 & 0.3144 & 0.1071 & 0.4364 & 0.4873 & 0.1185 & 0.5166 & 0.6564 \\
& RAG & 0.1719 & 0.4785 & 0.3229 & 0.1820 & 0.4578 & 0.4345 & 0.1319 & 0.5281 & 0.6640 \\
& PAG & 0.1816 & 0.4740 & 0.3130 & 0.2016 & 0.4517 & 0.4231 & 0.1292 & 0.5172 & 0.6537 \\
& Non-Think & 0.1295 & 0.4596 & 0.2973 & 0.1129 & 0.4341 & 0.4867 & 0.1217 & 0.5110 & 0.6546 \\
& RAG-Think & 0.1623 & 0.4695 & 0.3247 & 0.1564 & 0.4431 & 0.4498 & 0.1212 & 0.5154 & 0.6540 \\
& PAG-Think & 0.1665 & 0.4630 & 0.3079 & 0.1699 & 0.4398 & 0.4385 & 0.1296 & 0.5075 & 0.6490 \\
& R2P & 0.1786 & 0.4856 & 0.3289 & 0.1455 & 0.4558 & 0.5115 & 0.1362 & 0.5324 & 0.6664 \\
\midrule
\multirow{2}{*}{SFT} & SFT & 0.2691 & 0.5369 & 0.4372 & 0.2250 & 0.5064 & 0.5756 & 0.1700 & 0.5647 & 0.6816 \\
& NextQuill & 0.2689 & \textbf{0.5531} & \textbf{0.4443} & 0.2000 & 0.4998 & 0.5731 & 0.1709 & 0.5694 & \textbf{0.6909} \\
\midrule
\multirow{3}{*}{RL} & GRPO & 0.3476 & 0.5451 & 0.4035 & 0.2773 & 0.4959 & 0.5078 & 0.1600 & 0.5364 & 0.6792 \\
& GSPO & 0.3451 & 0.5418 & 0.3956 & 0.2743 & 0.4836 & 0.5664 & 0.1518 & 0.5395 & 0.6765 \\
& PrLM & 0.3469 & 0.5364 & 0.4004 & 0.2750 & 0.4698 & 0.4397 & 0.1620 & 0.5474 & 0.6893 \\
\midrule
\multirow{2}{*}{Hybrid} & TagPR & 0.3505 & 0.5410 & 0.3907 & 0.2748 & 0.4816 & 0.5578 & 0.1709 & 0.5499 & 0.6847 \\
& SFT+GRPO & \textbf{0.3523} & 0.5491 & 0.4277 & \textbf{0.2920} & \textbf{0.5166} & \textbf{0.6016} & \textbf{0.1749} & \textbf{0.5722} & 0.6887 \\
\bottomrule
\end{tabular*}
\end{table*}

\begin{table*}[htbp]
\centering
\caption{The prompt used to induce personalized user rubrics for the \textit{Amazon Review} dataset.}
\vspace{-3mm}
\fontsize{8pt}{8.5pt}\selectfont
\begin{tabularx}{\textwidth}{X}
\toprule
\textbf{Rubrics Induction - \textit{Amazon Review}} \\ \midrule
\textcolor{blue}{\textbf{[Instruction]}} \\
\raggedright\arraybackslash
\# Role \\
You are an expert in Personalization Analysis and Rubric Generation. Your job is to generate a self-contained set of personalized evaluation rubrics based on a user's historical reviews. These rubrics should be used to evaluate whether a newly generated review text matches the user's preferences. \\
\medskip
\# Input Data \\
- **User History:** A collection of historical review texts written by the target user. \\
\medskip
\# Task \\
Analyze the user's historical reviews from multiple dimensions (e.g., writing style, linguistic features, structure, tone, content focus, expression patterns) and generate 10 highly personalized rubrics that cover different aspects of the user's preferences. \\
\medskip
\# Drafting Rules \\
- **Atomicity:** Each rubric should be a single, clear and self-contained criterion that can be used to evaluate a review text independently. It should not require context from other rubrics or external information to be meaningful. Avoid combining multiple criteria into one rubric, as this makes evaluation ambiguous and reduces clarity. \\
- **Consistency**: Each rubric should reflect a consistent pattern identified across all the user's historical reviews. Each rubric must be satisfied when applied to all reviews in the user's profile. Avoid rubrics that only match a subset of the user's historical reviews, as they do not represent consistent patterns across the user's preferences. \\
- **Determinism:** Each rubric should be unambiguous and objectively evaluable, allowing for clear yes/no judgments rather than vague or subjective assessments. When applied to a review text, evaluators should be able to determine whether the rubric is satisfied without relying on personal interpretation or probabilistic reasoning. \\
- **Specificity:** Each rubric should be specific to a concrete, observable feature. Instead of vague statements like "matches writing style", provide precise characteristics such as "uses short, fragmented sentences without punctuation" or "prefers active voice over passive voice". \\ 
- **Item-Agnosticism:** Each rubric should be applicable to *any* future review of *any* item, not just the specific items in the user's history. Focus on *how* the user writes reviews (their writing style, structure, tone, linguistic patterns) rather than *what* items they reviewed (specific items, events, or subject matter). The rubrics should capture transferable writing characteristics that persist across different review contexts. \\
- **Multi-Dimensionality:** The set of rubrics should collectively cover multiple dimensions of the user's writing characteristics, such as writing style (sentence structure, paragraph organization), linguistic features (vocabulary choice, grammar patterns), tone (formality, emotional expression), content focus (what aspects are emphasized), and expression patterns (how ideas are conveyed). Avoid generating rubrics that all focus on the same dimension; instead, ensure diversity across different aspects of the user's preferences. \\
- **Personalization:** Each rubric should be personalized to the target user's historical reviews, capturing their unique writing characteristics and preferences. Each rubric should reflect something distinctive about this specific user's writing style that distinguishes them from other users. Generic rubrics that could apply to any user should be avoided; the rubrics must be tailored to reveal this particular user's preferences. \\
\medskip
\# Output Format \\
Example: \\
{[} \\
    \hspace*{4mm}\{"rule": "The review uses long, complex sentences with multiple clauses and sophisticated vocabulary, avoiding simplistic or fragmented phrasing."\}, \\
    \hspace*{4mm}\{"rule": "The review maintains a formal, essay-like tone with structured paragraphs, each focusing on a distinct analytical point or thematic thread."\}, \\
    \hspace*{4mm}\{"rule": "The review is structured with clearly labeled sections using uppercase headings followed by colons (e.g., 'THE STORY:', 'THE COOL THINGS:', 'BEST SCENES:', 'THE VERDICT:', 'GRADES:')."\}, \\
    \hspace*{4mm}\{"rule": "The structure follows a discursive, essay-like format with clear thematic paragraphs, each advancing a distinct analytical point without rigid section headings."\}, \\
    \hspace*{4mm}\{"rule": "The language includes informal but articulate expressions of personal judgment (e.g., 'wears thin VERY quickly', 'cool stuff!!!', 'decent as always') that convey strong subjective opinion without overly casual slang."\}, \\
    \hspace*{4mm}\{"rule": "The review employs a formal yet conversational tone with complex sentence structures, minimal contractions, and careful attention to grammar and punctuation."\}, \\
    \hspace*{4mm}\{"rule": "The review must contain a paragraph focused on the lead and supporting cast, highlighting the actors' performances with specific character names and roles, and often comparing the actor to their other well-known parts or assessing how they embody their character's moral or psychological complexity."\}, \\
    \hspace*{4mm}\{"rule": "The review explicitly mentions the source of the complimentary book copy (e.g., 'I received a complimentary copy from...') and specifies the publisher and review context (e.g., blog tour, NetGalley)."\}, \\
    \hspace*{4mm}\{"rule": "The review ends with a clear recommendation statement using the phrase 'I will recommend this book to others' or a close variant (e.g., 'will certainly recommend it')."\}, \\
    \hspace*{4mm}\{"rule": "The review concludes with a reflective, often uplifting or contemplative statement that underscores the lasting impact or significance of the work."\}, \\
{]} \\
\\
Provide a JSON array of objects. Each object must contain exactly one key: \texttt{rule}. \\
Do not output markdown code blocks or extra text. \\
\medskip
\textcolor{blue}{\textbf{[Inputs]}} \\
Given the following user historical reviews, please generate a self-contained set of personalized evaluation rubrics for the user. \\
<User Historical Reviews> \\
\{historical\_reviews\} \\
</User Historical Reviews> \\
\bottomrule
\end{tabularx}
\label{prompt_generate_amazon}
\end{table*}

\begin{table*}[htbp]
\centering
\caption{The prompt used to induce personalized user rubrics for the \textit{Reddit Post} dataset.}
\fontsize{8pt}{8.5pt}\selectfont
\vspace{-3mm}
\begin{tabularx}{\textwidth}{X}
\toprule
\textbf{Rubrics Induction - \textit{Reddit Post}} \\ \midrule
\textcolor{blue}{\textbf{[Instruction]}} \\
\raggedright\arraybackslash
\# Role \\
You are an expert in Personalization Analysis and Rubric Generation. Your job is to generate a self-contained set of personalized evaluation rubrics based on a user's historical reddit posts. These rubrics should be used to evaluate whether a newly generated reddit post content matches the user's preferences. \\
\medskip
\# Input Data \\
- **User History:** A collection of historical reddit post contents written by the target user. \\
\medskip
\# Task \\
Analyze the user's historical reddit posts from multiple dimensions (e.g., writing style, linguistic features, structure, tone, content focus, expression patterns) and generate 10 highly personalized rubrics that cover different aspects of the user's preferences. \\
\medskip
\# Drafting Rules \\
- **Atomicity:** Each rubric should be a single, clear and self-contained criterion that can be used to evaluate a reddit post content independently. It should not require context from other rubrics or external information to be meaningful. Avoid combining multiple criteria into one rubric, as this makes evaluation ambiguous and reduces clarity. \\
- **Consistency**: Each rubric should reflect a consistent pattern identified across all the user's historical reddit posts. Each rubric must be satisfied when applied to all reddit post contents in the user's profile. Avoid rubrics that only match a subset of the user's historical reddit posts, as they do not represent consistent patterns across the user's preferences. \\
- **Determinism:** Each rubric should be unambiguous and objectively evaluable, allowing for clear yes/no judgments rather than vague or subjective assessments. When applied to a reddit post content, evaluators should be able to determine whether the rubric is satisfied without relying on personal interpretation or probabilistic reasoning. \\
- **Specificity:** Each rubric should be specific to a concrete, observable feature. Instead of vague statements like "matches writing style", provide precise characteristics such as "uses short, fragmented sentences without punctuation" or "prefers active voice over passive voice". \\
- **Topic-Agnosticism:** Each rubric should be applicable to *any* future post of *any* topic, not just the specific posts in the user's history. Focus on *how* the user writes posts (their writing style, structure, tone, linguistic patterns) rather than *what* topics they posted (specific topics, events, or subject matter). The rubrics should capture transferable writing characteristics that persist across different post contexts. \\
- **Multi-Dimensionality:** The set of rubrics should collectively cover multiple dimensions of the user's writing characteristics, such as writing style (sentence structure, paragraph organization), linguistic features (vocabulary choice, grammar patterns), tone (formality, emotional expression), content focus (what aspects are emphasized), and expression patterns (how ideas are conveyed). Avoid generating rubrics that all focus on the same dimension; instead, ensure diversity across different aspects of the user's preferences. \\
- **Personalization:** Each rubric should be personalized to the target user's historical reddit post contents, capturing their unique writing characteristics and preferences. Each rubric should reflect something distinctive about this specific user's writing style that distinguishes them from other users. Generic rubrics that could apply to any user should be avoided; the rubrics must be tailored to reveal this particular user's preferences. \\
\medskip
\# Output Format \\
Provide a JSON array of objects. Each object must contain exactly one key: \texttt{rule}. \\
Example: \\
{[} \\
    \hspace*{4mm}\{"rule": "The post begins with a personal disclosure or contextual backstory that establishes the user's current emotional state, situation, or relevant history before introducing the main topic."\}, \\
    \hspace*{4mm}\{"rule": "The writing uses lowercase letters at the beginning of sentences inconsistently, often starting sentences with lowercase 'i' for first-person pronouns and occasionally neglecting standard capitalization at the beginning of sentences."\}, \\
    \hspace*{4mm}\{"rule": "The sentence structure must favor complete, grammatically standard sentences with consistent punctuation, avoiding fragmented prose, bullet points, or meta-commentary within the story itself."\}, \\
    \hspace*{4mm}\{"rule": "The post structures thoughts in a stream-of-consciousness style with minimal paragraph breaks, allowing one idea to flow into the next without formal transitions."\}, \\
    \hspace*{4mm}\{"rule": "The post must present biomechanical or anatomical reasoning to justify training recommendations, explicitly connecting muscle structure or joint mechanics to exercise selection or technique."\}, \\
    \hspace*{4mm}\{"rule": "The post balances emotional content with analytical reasoning, expressing personal feelings without becoming overly sentimental or dramatic."\}, \\
    \hspace*{4mm}\{"rule": "Writes in a direct, no-nonsense tone with minimal embellishment, prioritizing clarity and factual accuracy over emotional expression or storytelling."\}, \\
    \hspace*{4mm}\{"rule": "The post uses a conversational and informal tone with frequent use of first-person perspective (e.g., 'I', 'my', 'me') to express personal experiences, opinions, or questions."\}, \\
    \hspace*{4mm}\{"rule": "Vocabulary is plainspoken and accessible, favoring common idioms and colloquial expressions over technical jargon or literary flourishes, even when discussing sensitive or complex topics."\}, \\
    \hspace*{4mm}\{"rule": "The post concludes with a minimal or absent summary statement, instead ending directly on a question, emotional plea, or unresolved tension, leaving resolution open to community input."\}, \\
{]} \\
\\
Do not output markdown code blocks or extra text. \\
\medskip
\textcolor{blue}{\textbf{[Inputs]}} \\
Given the following user historical reddit posts, please generate a self-contained set of personalized evaluation rubrics for the user. \\
<User Historical Reddit Posts> \\
\{historical\_reddit\_posts\} \\
</User Historical Reddit Posts> \\
\bottomrule
\end{tabularx}
\label{prompt_generate_reddit}
\end{table*}
\begin{table*}[htbp]
\centering
\caption{The prompt used to induce personalized user rubrics for the \textit{News Headline} dataset.}
\fontsize{8pt}{8.5pt}\selectfont
\vspace{-3mm}
\begin{tabularx}{\textwidth}{X}
\toprule
\textbf{Rubrics Induction - \textit{News Headline}} \\ \midrule
\textcolor{blue}{\textbf{[Instruction]}} \\
\raggedright\arraybackslash
\# Role \\
You are an expert in Personalization Analysis and Rubric Generation. Your job is to generate a self-contained set of personalized evaluation rubrics based on a user's historical news. These rubrics should be used to evaluate whether a newly generated news headline matches the user's preferences. \\
\medskip
\# Input Data \\
- **User History:** A collection of historical news contents and their corresponding news headlines written by the target user. \\
\medskip
\# Task \\
Analyze how user transforms news content into news headline and generate 10 highly personalized rubrics that cover different aspects of the user's preferences. \\
\medskip
\# Drafting Rules \\
- **Atomicity:** Each rubric should be a single, clear and self-contained criterion that can be used to evaluate a news headline independently. It should not require context from other rubrics or external information to be meaningful. Avoid combining multiple criteria into one rubric, as this makes evaluation ambiguous and reduces clarity. \\
- **Consistency**: Each rubric should reflect a consistent pattern identified across all the user's historical news headlines. Each rubric must be satisfied when applied to all news headlines in the user's profile. Avoid rubrics that only match a subset of the user's historical news headlines, as they do not represent consistent patterns across the user's preferences. \\
- **Determinism:** Each rubric should be unambiguous and objectively evaluable, allowing for clear yes/no judgments rather than vague or subjective assessments. When applied to a news headline, evaluators should be able to determine whether the rubric is satisfied without relying on personal interpretation or probabilistic reasoning. \\
- **Specificity:** Each rubric should be specific to a concrete, observable feature. Instead of vague statements like "matches writing style", provide precise characteristics such as "uses short, fragmented sentences without punctuation" or "prefers active voice over passive voice". \\
- **News-Agnosticism:** Each rubric should be applicable to *any* future headline of *any* news, not just the specific news in the user's history. Focus on *how* the user writes news headline (their writing style, structure, tone, linguistic patterns) rather than *what* items they reviewed (specific items, events, or subject matter). The rubrics should capture transferable writing characteristics that persist across different news contexts. \\
- **Multi-Dimensionality:** The set of rubrics should collectively cover multiple dimensions of the user's writing characteristics, such as writing style (sentence structure, paragraph organization), linguistic features (vocabulary choice, grammar patterns), tone (formality, emotional expression), content focus (what aspects are emphasized), and expression patterns (how ideas are conveyed). Avoid generating rubrics that all focus on the same dimension; instead, ensure diversity across different aspects of the user's preferences. \\
- **Personalization:** Each rubric should be personalized to the target user's historical news, capturing their unique writing characteristics and preferences. Each rubric should reflect something distinctive about this specific user's writing style that distinguishes them from other users. Generic rubrics that could apply to any user should be avoided; the rubrics must be tailored to reveal this particular user's preferences. \\
\medskip
\# Output Format \\
Provide a JSON array of objects. Each object must contain exactly one key: \texttt{rule}. \\
Do not output markdown code blocks or extra text. \\
\medskip
\textcolor{blue}{\textbf{[Inputs]}} \\
Given the following user historical news, please generate a self-contained set of personalized evaluation rubrics for the user. \\
<User Historical News> \\
\{historical\_news\} \\
</User Historical News> \\
\bottomrule
\end{tabularx}
\label{prompt_generate_news}
\end{table*}

\begin{table*}[htbp]
\centering
\caption{The prompt used to evaluate the model-generated outputs and user-authored ground truth using personalized user rubrics on \textit{Amazon Review} dataset.}
\vspace{-3mm}
\fontsize{8pt}{9.2pt}\selectfont
\begin{tabularx}{\textwidth}{X}
\toprule
\textbf{Rubrics Evaluation - \textit{Amazon Review}} \\ \midrule
\textcolor{blue}{\textbf{[Instruction]}} \\
\raggedright\arraybackslash
\# Role \\
You are an expert Personalization Evaluator. Your job is to evaluate the quality of a review text based on a personalized evaluation rubric for the user. Given the metadata of the item and the review to be evaluated, the review text, and a personalized evaluation rubric, please judge whether the review text satisfies the given personalized evaluation rubric. \\
\vspace{1pt}
\# Input Data \\
- **Review Metadata:** The metadata of the item and the review to be evaluated. \\
- **Review Text:** The review text to be evaluated. \\
- **Personalized Evaluation Rubric:** The personalized evaluation rubric for the user to evaluate the review text. \\
\vspace{1pt}
\# Task \\
Judge whether the review text satisfies the given personalized evaluation rubric. \\
\vspace{1pt}
\# Output Format \\
Only output the answer of whether the review text satisfies the given personalized evaluation rubric. The answer should be \texttt{yes} or \texttt{no}. \\
\vspace{1pt}
\textcolor{blue}{\textbf{[Inputs]}} \\
<Review Metadata> \\
\{metadata\} \\
</Review Metadata> \\
\vspace{1pt}
<Review Text> \\
\{response\} \\
</Review Text> \\
\vspace{1pt}
<Personalized Evaluation Rubric> \\
\{rubric\} \\
</Personalized Evaluation Rubric> \\
\bottomrule
\end{tabularx}
\label{prompt_evaluate_amazon}
\end{table*}

\begin{table*}[htbp]
\centering
\caption{The prompt used to evaluate the model-generated outputs and user-authored ground truth using personalized user rubrics on \textit{Reddit Post} dataset.}
\vspace{-3mm}
\fontsize{8pt}{9.2pt}\selectfont
\begin{tabularx}{\textwidth}{X}
\toprule
\textbf{Rubrics Evaluation - \textit{Reddit Post}} \\ \midrule
\textcolor{blue}{\textbf{[Instruction]}} \\
\raggedright\arraybackslash
\# Role \\
You are an expert Personalization Evaluator. Your job is to evaluate the quality of a reddit post content based on a personalized evaluation rubric for the user. Given the reddit post summary, the reddit post content to be evaluated, and a personalized evaluation rubric, please judge whether the reddit post content satisfies the given personalized evaluation rubric. \\
\vspace{1pt}
\# Input Data \\
- **Reddit Post Summary:** The summary of the reddit post. \\
- **Reddit Post Content:** The content of the reddit post to be evaluated. \\
- **Personalized Evaluation Rubric:** The personalized evaluation rubric for the user to evaluate the reddit post content. \\
\vspace{1pt}
\# Task \\
Judge whether the reddit post content satisfies the given personalized evaluation rubric. \\
\vspace{1pt}
\# Output Format \\
Only output the answer of whether the reddit post content satisfies the given personalized evaluation rubric. The answer should be \texttt{yes} or \texttt{no}. \\
\vspace{1pt}
\textcolor{blue}{\textbf{[Inputs]}} \\
<Reddit Post Summary> \\
\{summary\} \\
</Reddit Post Summary> \\
\vspace{1pt}
<Reddit Post Content> \\
\{content\} \\
</Reddit Post Content> \\
\vspace{1pt}
<Personalized Evaluation Rubric> \\
\{rubric\} \\
</Personalized Evaluation Rubric> \\
\bottomrule
\end{tabularx}
\label{prompt_evaluate_topic}
\end{table*}

\begin{table*}[htbp]
\centering
\caption{The prompt used to evaluate the model-generated outputs and user-authored ground truth using personalized user rubrics on \textit{News Headline} dataset.}
\vspace{-3mm}
\fontsize{8pt}{9.2pt}\selectfont
\begin{tabularx}{\textwidth}{X}
\toprule
\textbf{Rubrics Evaluation - \textit{News Headline}} \\ \midrule
\textcolor{blue}{\textbf{[Instruction]}} \\
\raggedright\arraybackslash
\# Role \\
You are an expert Personalization Evaluator. Your job is to evaluate the quality of a news headline based on a personalized evaluation rubric for the user. Given the news content, the news headline to be evaluated, and a personalized evaluation rubric, please judge whether the news headline satisfies the given personalized evaluation rubric. \\
\vspace{1pt}
\# Input Data \\
- **News Content:** The content of the news. \\
- **News Headline:** The news headline to be evaluated. \\
- **Personalized Evaluation Rubric:** The personalized evaluation rubric for the user to evaluate the news headline. \\
\vspace{1pt}
\# Task \\
Judge whether the news headline satisfies the given personalized evaluation rubric. \\
\vspace{1pt}
\# Output Format \\
Only output the answer of whether the news headline satisfies the given personalized evaluation rubric. The answer should be \texttt{yes} or \texttt{no}. \\
\vspace{1pt}
\textcolor{blue}{\textbf{[Inputs]}} \\
<News Content> \\
\{content\} \\
</News Content> \\
\vspace{1pt}
<News Headline> \\
\{response\} \\
</News Headline> \\
\vspace{1pt}
<Personalized Evaluation Rubric> \\
\{rubric\} \\
</Personalized Evaluation Rubric> \\
\bottomrule
\end{tabularx}
\label{prompt_evaluate_news}
\end{table*}

\begin{table*}[htbp]
\centering
\caption{The prompt used to evaluate the model-generated outputs and user-authored ground truth using LLM-as-a-judge on \textit{Amazon Review} dataset.}
\vspace{-3mm}
\fontsize{8pt}{9.2pt}\selectfont
\begin{tabularx}{\textwidth}{X}
\toprule
\textbf{LLM-as-a-judge Evaluation - \textit{Amazon Review}} \\ \midrule
\textcolor{blue}{\textbf{[Instruction]}} \\
\raggedright\arraybackslash
\# Role \\
You are an expert Personalization Evaluator. Your job is to evaluate the quality of a review text based on the user histories. Given the user historical reviews, the metadata of the item and the review to be evaluated, the review text, please judge how well the review text matches the user's preferences. \\
\vspace{1pt}
\# Input Data \\
- **User History:** A collection of historical review texts written by the target user. \\
- **Review Metadata:** The metadata of the item and the review to be evaluated. \\
- **Review Text:** The review text to be evaluated. \\
\vspace{1pt}
\# Output Format \\
Only output the score of the review text. The score should be a number between 1 and 5. \\
Do not output markdown code blocks or extra text. \\
\vspace{1pt}
\textcolor{blue}{\textbf{[Inputs]}} \\
<User Historical Reviews> \\
\{past\_reviews\} \\
</User Historical Reviews> \\
\vspace{1pt}
<Review Metadata> \\
\{meta\_info\} \\
</Review Metadata> \\
\vspace{1pt}
<Review Text> \\
\{review\} \\
</Review Text> \\
\bottomrule
\end{tabularx}
\label{prompt_llm_as_judge_amazon}
\end{table*}

\begin{table*}[htbp]
\centering
\caption{The prompt used to evaluate the model-generated outputs and user-authored ground truth using LLM-as-a-judge on \textit{Reddit Post} dataset.}
\vspace{-3mm}
\fontsize{8pt}{9.2pt}\selectfont
\begin{tabularx}{\textwidth}{X}
\toprule
\textbf{LLM-as-a-judge Evaluation - \textit{Reddit Post}} \\ \midrule
\textcolor{blue}{\textbf{[Instruction]}} \\
\raggedright\arraybackslash
\# Role \\
You are an expert Personalization Evaluator. Your job is to evaluate the quality of a reddit content based on the user histories. Given the user historical reddit post contents, the reddit post summary, and the reddit post content to be evaluated, please judge how well the reddit content matches the user's preferences. \\
\vspace{1pt}
\# Input Data \\
- **User History:** A collection of historical reddit post contents written by the target user. \\
- **Reddit Post Summary:** The summary of the reddit post. \\
- **Reddit Post Content:** The content of the reddit post to be evaluated. \\
\vspace{1pt}
\# Output Format \\
Only output the score of the reddit content. The score should be a number between 1 and 5. \\
Do not output markdown code blocks or extra text. \\
\vspace{1pt}
\textcolor{blue}{\textbf{[Inputs]}} \\
<User Historical Reddit Posts> \\
\{past\_posts\} \\
</User Historical Reddit Posts> \\
\vspace{1pt}
<Reddit Post Summary> \\
\{summary\} \\
</Reddit Post Summary> \\
\vspace{1pt}
<Reddit Post Content> \\
\{content\} \\
</Reddit Post Content> \\
\bottomrule
\end{tabularx}
\label{prompt_llm_as_judge_topic}
\end{table*}

\begin{table*}[htbp]
\centering
\caption{The prompt used to evaluate the model-generated outputs and user-authored ground truth using LLM-as-a-judge on \textit{News Headline} dataset.}
\vspace{-3mm}
\fontsize{8pt}{9.2pt}\selectfont
\begin{tabularx}{\textwidth}{X}
\toprule
\textbf{LLM-as-a-judge Evaluation - \textit{News Headline}} \\ \midrule
\textcolor{blue}{\textbf{[Instruction]}} \\
\raggedright\arraybackslash
\# Role \\
You are an expert Personalization Evaluator. Your job is to evaluate the quality of a news headline based on the user histories. Given the user historical news, the news content, and the news headline to be evaluated, please judge how well the news headline matches the user's preferences. \\
\vspace{1pt}
\# Input Data \\
- **User History:** A collection of historical news contents and their corresponding news headlines written by the target user. \\
- **News Content:** The content of the news. \\
- **News Headline:** The news headline to be evaluated. \\
\vspace{1pt}
\# Output Format \\
Only output the score of the news headline. The score should be a number between 1 and 5. \\
Do not output markdown code blocks or extra text. \\
\vspace{1pt}
\textcolor{blue}{\textbf{[Inputs]}} \\
<User Historical News> \\
\{past\_news\} \\
</User Historical News> \\
\vspace{1pt}
<News Content> \\
\{content\} \\
</News Content> \\
\vspace{1pt}
<News Headline> \\
\{headline\} \\
</News Headline> \\
\bottomrule
\end{tabularx}
\label{prompt_llm_as_judge_news}
\end{table*}

\end{document}